\documentclass[11pt]{elsarticle}
\usepackage{lineno,hyperref}
\modulolinenumbers[5]

\journal{Preprint}

\usepackage[margin=1.25in]{geometry}
\usepackage{amsthm}
\usepackage{amsmath}
\usepackage{amsfonts}
\usepackage[percent]{overpic}
\usepackage{graphicx,graphics,epsfig}
\usepackage{amsmath,amssymb,dsfont}
\usepackage{mathrsfs}
\usepackage{arydshln}
\usepackage{threeparttable}
\usepackage{booktabs}
\usepackage{multirow}
\usepackage{enumerate}
\usepackage[usenames,dvipsnames]{color}
\usepackage{subfigure}
\usepackage[normalem]{ulem}
\usepackage{bm}
\usepackage{upgreek}
\usepackage{url}
\usepackage{color}
\usepackage{setspace}
\usepackage{dsfont}

\usepackage{algorithm}
\usepackage{algpseudocode}

\begin{document}

\begin{frontmatter}
	
	\title{Adaptively trained Physics-informed Radial Basis Function Neural Networks for Solving Multi-asset Option Pricing Problems}
	    \author[a,b]{Yan Ma}
        \author[b]{Yumeng Ren \corref{mycorrespondingauthor}}
       
        \cortext[mycorrespondingauthor]{Corresponding author}
        \ead{ymren3-c@my.cityu.edu.hk}
        \author[c]{Elisabeth Larsson} 
        
        \address[a]{College of Mathematics and Statistics, Northwest Normal University, Lanzhou, China}
        \address[b]{Department of Mathematics, City University of Hong Kong, Hong Kong SAR, China}
        \address[c]{Department of Information Technology, Uppsala University, Sweden}

\begin{abstract}
The present study investigates the numerical solution of Black-Scholes partial differential equation (PDE) for option valuation with multiple underlying assets. We develop a physics-informed (PI) machine learning algorithm based on a radial basis function neural network (RBFNN) that concurrently optimizes the network architecture and predicts the target option price. The physics-informed radial basis function neural network (PIRBFNN) combines the strengths of the traditional radial basis function collocation method and the physics-informed neural network machine learning approach to effectively solve PDE problems in the financial context. By employing a PDE residual-based technique to adaptively refine the distribution of hidden neurons during the training process, the PIRBFNN facilitates accurate and efficient handling of multidimensional option pricing models featuring non-smooth payoff conditions. The validity of the proposed method is demonstrated through a set of experiments encompassing a single-asset European put option, a double-asset exchange option, and a four-asset basket call option.
\end{abstract}

	
\end{frontmatter}

\section{Introduction}
Financial markets, where assets such as stocks, bonds, foreign exchange, and physical commodities are traded, make up an increasing share of the global economy. Their stability is therefore of the utmost importance.  Financial derivatives are contracts involving the buying, selling, or swapping  of one or more assets that can be used for risk mitigation, also called hedging. The most prevalent forms of financial derivatives are the so called options~\cite{Paul1995Options}.   
For the market to function well, these option contracts need to be priced accurately, such that no arbitrage opportunities are introduced. They also need to be priced in real time when a trade takes place. In most cases closed form solutions are not available, and pricing has to be carried out using numerical methods. The most straightforward method is to perform Monte Carlo simulation based on the stochastic diffusion processes describing the underlying asset prices~\cite{Glasserman}. This scales linearly with the number of underlying assets, but typically requires a large number of simulated trajectories to reach a small enough statistical error in the option price.

For a low to moderate number of underlying assets, it can be more efficient to solve the Black-Scholes partial differential equation (PDE)~\cite{WOS:000363753800003}. Existing literature contains a variety of numerical techniques for solving both one-dimensional and multi-dimensional Black-Scholes equations. Among the mesh-based methods employed within this field of study, finite difference methods (FDM) \cite{WOS:000679358400002, WOS:000273521800008, WOS:000524085900090, WOS:000753857500013} and finite element methods (FEM) \cite{, WOS:000334824500004, WOS:000305360000022, WOS:000365740300017} are the two predominant numerical discretization approaches. For FDM, mesh refinement involves whole gridlines leading to unnecessary refinment far away from the region of interest and increased computational cost, while for FEM, local refinement is available, but may be non-trivial to implement in more than three dimensions. As an alternative, various meshless methods have been proposed as competitive solvers for financial applications \cite{WOS:000371837600015, WOS:000526486500001, WOS:001148177600001, WOS:000462172900005}.

An advantage of meshless methods is the simplicity of the solution process. These methods work directly with scattered point clouds that can be generated and locally refined in any number of dimensions. Within this category of methodology, the radial basis function (RBF) methods have been shown to perform well both for function approximation and solution of PDEs \cite{WOS:000363753800003, PETTERSSON200882, SHCHERBAKOV2016185, WOS:000457316600035, WOS:000514017700015}. When solving a boundary value PDE problem using the traditional RBF method, an RBF approximation $\hat{V}(x)$ to the unknown function $V(x)$ has the general form
\begin{equation}
  \hat{V}(x)=\sum_{j}W_j\phi(C_j\|x-x_j\|) + P(x),
\end{equation}
where $W_j$ are unknown weights to determine, $\phi(r)$ is a radial basis kernel function, $x_j\in\mathbb{R}^d$ are centre points, $C_j$ are called shape parameters, and $P(x)$ is a low order polynomial also involving unknown weights. The kernel function, shape parameters, and centre locations need to be specified prior to the solution process, and then the weights (coefficients) can be determined through collocation with given data, resulting in a linear system to solve. Once the weights are computed, the approximation can be calculated at any point within the domain. Although the aforementioned solving process may appear straightforward, a challenge is the selection of the problem parameters, and in particular, determining the optimal shape parameter values is an open research problem.  

Driven by the proliferation of accessible data and computing resources, machine learning algorithms have gained significant interest in many areas of research. The field of scientific computing has, e.g., witnessed widespread utilization of deep neural networks for solving PDEs. Among these approaches, the physics-informed neural network (PINN) stands out as the particularly promising one. By leveraging the power of automatic differentiation technique which is conveniently implemented in the popular machine learning libraries like TensorFlow and PyTorch, PINN incorporates both PDE constraints and boundary measurements into the loss function and minimizes the loss to train the neural network. The key distinguishing feature of PINN lies in its ability to encode the physical laws directly into a learning system. This integration enhances the information content of the data that the algorithm sees, enabling PINN to quickly converge towards the right solution while also exhibiting strong generalization capability
\cite{WOS:000453776000028, WOS:000423263500003, WOS:000638011900006, WOS:000793366500003, WOS:000955122000001, WOS:000805057700001}.
Despite the historical successes in tackling a wide range of PDE problems with varying complexities, the architecture of deeper layers combined with multiple hidden neurons necessitates increased feeding resources for effective training and computational efforts to open up the possibility of using sophisticated second-order gradient decent optimizer.

In this study, we combine the machine learning dynamics of the PINN with the approximation power of the RBFs, effectively  overcoming the limitations inherent in each approach. Specifically speaking, we employ RBFs as activation functions to construct novel physics-informed radial basis function neural networks (PIRBFNNs) for pricing European options in one, two, and four space dimensions.  
See~\cite{WOS:001158937800001} for similar type of method, applied to other types of PDE problems. Compared to the traditional RBF collocation method, the RBFNN can simultaneously provide accurate option prices and learn appropriate RBF method parameters. The mechanism of machine learning based on artificial neuron network empowers RBFNN to preserve the characteristic of meshless computation and have the potential to surmount the curse of dimensionality. The solution to the PDE problem materializes gradually through the training process, underscoring the necessity for the development of fast learning algorithms. Several early studies have delved into the analysis of various gradient algorithms for learning RBFNNs, including the two-stage strategy \cite{WOS:000184011900027}, trust region method \cite{WOS:000394351900012}, and Levenberg-Marquardt method \cite{Filippov:2019, Vladimir:2019, Gorbachenko:2019}.
The two-stage learning process entailed sequential optimization of weights and RBF parameters, rather than concurrent refinement. While in the PIRBFNN, we employ simultaneous training of centre locations, shape parameters, and linear weights. Although both the trust region and Levenberg-Marquardt methods offer rapid second-order optimization, the former is complicated to implement, and the latter requires evaluating the conditioning of the system being solved. In both scenarios, achieving a computational time reduction is possible by providing the exact Jacobian of the error function vector. However, this calculation becomes nontrivial in high-dimensional spaces. Compared to the PINN, the PIRBFNN employs a shallow neural network architecture instead of the multi-layered structure, significantly streamlining the backpropagation process of automatic differentiation. Advanced second-order algorithms like Limited Memory Broyden-Fletcher-Goldfarb-Shanno (L-BFGS) can be extended readily to the optimization of PIRBFNN, while also can be conveniently implemented by using the optimizer package within Python's working environment. Despite lacking deep data mining capabilities, we augment the information of the hidden layer RBF neurons by considering both the shape parameter values and centre spatial locations as dynamically trainable parameters to enhance the representative ability of the PIRBFNN. As documented in the literature, the number of RBF neurons typically remains constant throughout the network training process \cite{Gorbachenko:2022, WOS:000874618900004, WOS:001240386200001}. Conversely, our learning scheme introduces a residual-based refinement strategy that adaptively inserts additional RBF neurons during training at locations where the PDE residual is significant \cite{WOS:000615712100007}. By leveraging this adaptive strategy to optimize the network structure, we can enhance the training dynamic system and assist the optimizer to increase attention to the challenging regions of information propagation, thereby boosting prediction accuracy while accelerating training convergence.   

The remainder of the paper is organized as follows: Section 2 gives a succinct description on the general form of the multivariate Black-Scholes model for pricing a European  option involving several assets. In Section 3, the proposed PIRBFNN architecture and adaptive training dynamics to tackle a terminal-boundary value problem in the financial context are presented. Section 4 is dedicated to the numerical implementation of three representative option pricing models with different governing equations, boundary conditions, and dimensions to verify the feasibility of the proposed method. Finally, in Section 5 we provide some concluding remarks along with future research directions.

\section{Black-Scholes model for multi-asset option pricing problem}
The widely recognized Black-Scholes equation, which models the pricing of single-asset and multi-asset European style options, is generally expressed as follows:
\begin{equation}
	 \label{BS_equation_general}
		\dfrac{\partial V(\bm{S},t)}{\partial t} + \frac{1}{2} \sum_{i,j=1}^{d}\rho_{ij} \sigma_i \sigma_j S_i S_j \dfrac{\partial^2 V(\bm{S},t)}{\partial S_i \partial S_j} +  \sum_{i=1}^{d} r S_i \dfrac{\partial V(\bm{S},t)}{\partial S_i} -r V(\bm{S},t) = 0,
\end{equation}
where $\bm{S} = (S_1,S_2,\dots,S_d) \in [0,\infty)^d$ is the vector of underlying assets prices $S_1,S_2,\dots,S_d$, time $t \in [0,T]$, where $T$ is the expiration time of the option, $V(\bm{S},t)$ is the value of the option, $\sigma_i$ is the volatility of the $i$-th asset, $r$ is the risk-free interest rate,  and $\rho_{ij}$ is the correlation between the $i$-th asset and the $j$-th asset. The correlation matrix must satisfy $\rho_{ii} = 1$, $\rho_{ij} = \rho_{ji}$, $i \ne j$, and $\left| \rho_{ij} \right| \leq 1$.

The terminal condition for this backward Eq.\eqref{BS_equation_general} is determined by the payoff value of the option at maturity, that is
\begin{equation}
	 \label{BS_payoff_general}
	  V(\bm S,T) = V_T(\bm S),
\end{equation}
where the so called payoff function $V_T$ varies across different types of option contracts. 

The PDE \eqref{BS_equation_general} is a second-order parabolic equation, that requires appropriate boundary conditions for well-posedness. For a discussion of the requirements on boundary conditions for this type of problems, see~\cite{Tysk,Ficchera}. In particular, we impose spatial boundary conditions at near-field boundaries
$(S_i = 0,i = 1,2,...,d)$ and far-field boundaries ($S_i = \infty,i = 1,2,...,d)$. It should be noted that the spatial domain is semi-infinite, for numerical implementation it needs to be truncated to a bounded version. We define the spatial computational domain as $\Omega = (0, S_{max})^d$, where the value of $S_{\max}$ is normally selected to be several times larger than the strike price $K$ specified in the option contract, ensuring that negative effects on the computational accuracy of the option price $V(\bm S, t)$ due to the far-field boundary conditions are negligible. 

We summarize the numerical computation of the multi-asset European option pricing problem as solving the PDE 
\begin{equation}
	\label{Gov_equation_general}
	\mathcal{L} V(\bm S,t) = 0,~~~~~~~~S \in \Omega,~~ t \in [0,T)
\end{equation}
subject to the terminal time condition
\begin{equation}
	\label{Ter_condition_general}
	V(\bm S,T) = V_T(\bm S),~~~~~~S \in \Bar \Omega
\end{equation}
and the spatial boundary conditions
\begin{equation}
	\label{Bou_condition_general}
	\mathcal{B} V(\bm S,t)= g(\bm S,t),~~~~~~S \in \partial\Omega_\mathcal{B},~~t \in [0,T)
\end{equation}
where the linear differential operator $\mathcal{L} = \dfrac{\partial}{\partial t} + \frac{1}{2} \sum \limits_{i,j=1}^{d}\rho_{ij} \sigma_i \sigma_j S_i S_j \dfrac{\partial^2}{\partial S_i \partial S_j} + \sum \limits_{i=1}^{d} r S_i \dfrac{\partial}{\partial S_i} -r$, the problem dependent boundary condition operators are represented by $\mathcal{B}$ and are applied on the boundary of the spatial domain denoted by $\partial \Omega_\mathcal{B}$. The boundary function $g(\bm S,t)$ is prescribed problem dependent function.

The terminal-boundary value problem (TBVP) \eqref{Gov_equation_general}--\eqref{Bou_condition_general} for multi-asset option pricing presents a challenging computational task within the field of financial mathematics. Firstly, the dimension of the governing PDE is given by the number of underlying assets, leading to an exponential increase in the cost in terms of assets (the curse of dimensionality). Furthermore, closed-form solutions could not always be found for high-dimensional European options, with basket options being a prime example. In addition, artificial boundary conditions need to be introduced to reduce the infinite problem domain to a bounded computational domain. For some types of options accurate asymptotic boundary conditions are known, while for other cases a large domain may be needed to reduce effects from inaccurate boundary conditions. Lastly, the contract funtion (payoff) that serves as the final condition of the PDE typically involves non-smoothness in the value or in the derivatives that calls for adaptive treatment of that region. The low regularity in the terminal data can decrease the convergence order and the accuracy level. Investigating how the PIRBFNN performs for such financial models ranging from low-dimensional to moderately high-dimensional is the primary objective of this study. The coming section offers a detailed demonstration of our proposal.

\section{Physics-informed radial basis function neural network (PIRBFNN) for solving terminal-boundary value problem}

\subsection{Architecture of radial basis function neural network (RBFNN)}

We employ a RBFNN to generate an approximate solution $\hat V(\bm S,t)$ of the unknown option price $V(\bm S,t)$ for the TBVP \eqref{Gov_equation_general}--\eqref{Bou_condition_general}.

RBF methods for time-dependent PDEs as the Black-Scholes equation typically combine a spatial discretization with a time-stepping scheme. However, in the RBFNN, we adopt a space-time approach by treating the temporal variable $t$ as an additional spatial dimension. The terminal condition becomes a Dirichlet boundary condition applied only at the final time boundary. The TBVP \eqref{Gov_equation_general}--\eqref{Bou_condition_general} is in this way recast as a boundary value problem (BVP) within the spatial-temporal domain $\bar{\Omega}\times[0,T]$.

The RBFNN is a feedforward neural network comprising two layers. The input layer contains the coordinates of a training point $(\bm S,t) = (S_1,S_2,\dots,S_d,t)$ in the spatial-temporal domain, representing information about the spot prices of the underlying assets $(S_1,S_2,\dots,S_d)$ at the corresponding time $t$ fed into the network. The hidden layer performs a nonlinear transformation using the RBF $\phi$ as the activation function. Each RBF neuron is associated with a centre point $(\bar {\bm S}_n,\bar t_{d+1,n}) = (\bar S_{1n},\bar S_{2n},\dots,\bar S_{dn},\bar t_{d+1,n})$ and either a shape parameter scalar $C_n$ or a shape parameter vector $\bm C_n = (C_{1n},C_{2n},\dots,C_{dn},C_{d+1,n})$, where the components of $\bm C_n$ act as shape parameters for the different dimensions in the RBF kernel function.
The activation mapping is contingent upon the scaled Euclidean distance $\lVert \cdot \lVert_{\bm C}$ between the input training point $(\bm S,t)$ and the RBF centre point $(\bar{\bm S}_n,\bar t_n)$. We can write the activation function corresponding to the $n$-th neuron as
\begin{equation}
  \phi_n(\bm S,t) = \phi(\lVert (\bm S,t)-(\bar {\bm S}_n,\bar t_{d+1,n}) \rVert_{\bm C_n}),
\end{equation}
where $\lVert (\bm S,t)-(\bar {\bm S}_n,\bar t_n) \rVert_{\bm C_n}^2=C_{1n}^2(S_1-\bar{S}_{1n})^2+\cdots+C_{dn}^2(S_d-\bar{S}_{dn})^2+C_{d+1,n}^2(t-\bar t_{d+1,n})^2 $. For a scalar shape parameter, this is reduced to $\lVert\cdot \rVert_{C_n}^2=C_n^2\|\cdot\|^2$.
The predicted option price is determined by a linear weighted sum of form \eqref{approximant} from the output layer, which can be mathematically expressed as:

\begin{equation}\label{approximant}
	\hat V(\bm S,t) = \sum_{n=1}^{N} W_n \phi_n(\bm S,t)+W_{N+1},
\end{equation}
where $N$ is the number of hidden RBF neurons, $W_n$ is the weight of the $n$-th neuron, $W_{N+1}$ refers to the network bias.
Taking Gaussian RBF neurons as an example, the approximant~\eqref{approximant} can be explicitly written as follows:

\begin{equation*}
   \hat V(\bm S,t) = \sum_{n=1}^{N} W_n e^{- \left[C_{1n}^2(S_1-\bar{S}_{1n})^2+\cdots+C_{dn}^2(S_d-\bar{S}_{dn})^2+C_{d+1,n}^2(t-\bar t_{d+1,n})^2\right]}+W_{N+1}.
\end{equation*}

All the parameters mentioned in the above network approximation, including the number of RBF neurons $N$, the centre locations $L = \{(\bar {\bm S}_1,\bar t_1),(\bar {\bm S}_2,\bar t_2),\cdots,(\bar {\bm S}_N,\bar t_N)\}$, the shape parameters  $C = \{\bm C_1,\bm C_2,\cdots,\bm C_N\}$, and the linear weights $W = \{W_1,W_2,\cdots,W_N,W_{N+1}\}$ are trainable objects during the machine learning process. 

Remark: To enhance the RBFNN's representational capacity while avoiding overfitting, we adopt the scalar shape parameters for option pricing problem with a single underlying asset, and the vector form of the shape parameters for more challenging problems involving two or more underlying assets.

\subsection{Generation of the training points for various constraints}
To incorporate physics-informed constraints implied by the governing equation and boundary conditions into the training process of RBFNN, it is essential to construct a training points set. This set consists of multiple subsets of collocation points, each chosen from a designated region to assess the deviation of the neural network prediction from the corresponding constraints.

Specifically, by fixing the seed value in the random Python module, we employ the pseudo-random number generator to create a deterministic training points set $P = \{(\bm S_1,t_1),\\(\bm S_2,t_2),\cdots,(\bm S_M,t_M)\}$. These points are random samples from a uniform distribution across the entire spatial-temporal domain, including the interior and the boundaries. We partition $P$ into three subsets:
\begin{equation}
P = P_\mathcal{L} \bigcup P_{T} \bigcup P_{\mathcal{B}}, 
\end{equation}
where
\begin{equation*}
	\renewcommand{\arraystretch}{2}
	\begin{array}{l}
	P_{\mathcal{L}} = \{(\bm S_i,t_i)\}_{i=1}^{M_\mathcal{L}} \subset \Omega \times (0,T), \\
	P_{T} = \{(\bm S_i,t_i)\}_{i=M_\mathcal{L}+1}^{M_T} \subset \Omega \times \{T\}, \\
        P_{\mathcal{B}} = \{(\bm S_i,t_i)\}_{i=M_\mathcal{L}+M_T+1}^{M_\mathcal{L}+M_T+M_\mathcal{B}} \subset \partial \Omega_\mathcal{B} \times (0,T), \\
	\end{array}
\end{equation*}
$M=M_\mathcal{L}+M_T+M_\mathcal{B}$. These training point subsets correspond to the collocation points assigned for feeding into the PDE, the terminal time boundary condition, and the spatial boundary conditions, respectively.

\subsection{Formulation of the loss function}
To achieve an accurate neural network approximator $\hat V(\bm S,t)$ with well-tuned parameters $\theta = \{N;L,C,W\}$, it is essential to define an effective loss function that encapsulates the physics-informed residual errors derived from the governing equation and boundary conditions. In the current study, we utilize the conventional loss formulation of a PINN within the RBFNN framework, embedding the physical law regularization and observation data fitting into the loss function
\begin{equation}
	\ell (\theta;P) = \ell (\theta;P_{\mathcal{L}}) + \ell (\theta;P_{T}) + \ell (\theta;P_{\mathcal{B}}),
\end{equation}
where
\begin{equation*}
  \renewcommand{\arraystretch}{2}
  \begin{array}{l}
    \ell (\theta;P_\mathcal{L}) = \frac{1}{M_\mathcal{L}} \sum\limits_{(\bm S_i,t_i) \in P_\mathcal{L}} \lVert \mathcal{L} \hat V(\bm S_i,t_i) - f(\bm S_i,t_i) \lVert_2^2,  \\
    \ell (\theta;P_T) = \frac{1}{M_T} \sum\limits_{(\bm S_i,T) \in P_T} \lVert   \hat V(\bm S_i,T) - V_T(\bm S_i) \lVert_2^2,  \\
    \ell (\theta;P_\mathcal{B}) = \frac{1}{M_\mathcal{B}} \sum\limits_{(\bm S_i,t_i) \in P_{\mathcal{B}}} \lVert \mathcal B \hat V(\bm S_i,t_i) - g(\bm S_i,t_i) \lVert_2^2,
	\end{array}
\end{equation*}
where $f(\bm S_i,t_i)\equiv 0$ and $\lVert \cdot \lVert_2$ represents the $L_2$-norm. 

Remark: Based on extensive experiments with different values of the penalty parameters associated with the PDE loss and the boundary loss, it is observed that varying these parameters does not lead to noticeable improvements in either the prediction accuracy or the convergence speed of the network. To reduce the number of hyperparameters and simplify the training process, all penalty parameters corresponding to different loss terms are therefore set to $1$ in our loss function.

\subsection{Initialization of the network parameters}
To start the learning process of PIRBFNN, all the parameters $\theta = \{N;L,C,W\}$ defining the neural network should be given their initial settings. These values are then optimized automatically through a gradient descent algorithm combined with adaptive training dynamics. The parameter choice should be as automatic as possible and work for a large family of problems, but should also be a reasonable starting point for the optimizer. 

$N$ represents the number of the RBF neurons. Increasing the value of $N$ widens the hidden layer of the neural network, thereby improving the expressiveness of the RBFNN. Relative to a given amount of training points, we propose a PDE residual-based technique to adaptively refine the number of RBF hidden neurons from an initial value up to a specific limit. Consequently, the network architecture can be efficiently designed with an appropriate width. The detailed execution process will be described in the following subsection.

$L$ controls the centre locations of the RBF neurons. To sample the initial RBF centres uniformly within the computational domain and make them consistent across multiple re-runs of the program, the coordinates of the initial centre points are generated using the Python random module with a specified seed value. Subsequently, the newly added centres of RBF neurons are positioned at locations with significant PDE residual errors, as identified by the adaptive algorithm during the training process.

$C$ describes the shape characteristics of the RBF neurons. For the single-asset option pricing problem $(d=1)$, scalar shape parameters are used and the values $\{C_1, C_2,\cdots,C_N\}$ for the $N$ RBF neurons are initialized as the numbers uniformly distributed within the range $(0,1)$ utilizing the random number generator in PyTorch, with a seed argument for reproducibility. For the multi-asset option pricing problems $(d \geq 2)$, choosing a random initial shape parameter scalar for each RBF neuron did not lead to competetive results. Different functions for determining the initial shape parameters were evaluated and the following approach was selected:

For each RBF neuron with initial centre point $(\bar S_{1n}, \cdots, \bar S_{in}, \cdots, \bar S_{dn}, \bar t_{d+1,n})$, the initial component $C_{in}$ of the shape parameter vector $\bm C_n = (C_{1n}, \dots, C_{in}, \dots, C_{dn}, C_{d+1,n})$ is determined as
\begin{equation}
 C_{in} =
\begin{cases}
\frac{\alpha \cdot \sqrt{D_i Z_{\alpha}}}{\sqrt {(2 D_i Z_{\alpha}-\bar S_{in})^3+(\bar S_{in})^3}},~~~(i =1, 2, \dots, d),  
\\[15pt]
\frac{\alpha \cdot \sqrt{T_i Z_{\alpha}}}{\sqrt {(2 T_i Z_{\alpha}-\bar t_{in})^3+(\bar t_{in})^3}},~~~~~(i =d+1),
\end{cases}
\end{equation}
where $i=1,2,\cdots,d$ correspond to the spatial dimensional indexes, and $i = d+1$ corresponds to the temporal dimensional index. This formula holds for all $N$ neurons, $n = 1,2,\cdots,N$. $D_i = S_{\max}$ for $i=1,2,\cdots,d$ and $T_i = T$ for $i = d+1$. $Z_{\alpha} (0 < Z_{\alpha} \le 1)$ and $\alpha (\alpha > 0)$ are two parameters that respectively determine the peak location and the corresponding peak magnitude of the shape parameters $C = \{\bm C_1,\bm C_2,\cdots,\bm C_N\}$ in each dimension. In our experiments, $Z_{\alpha} = 0.5$ is chosen so that the shape parameters attain their maximum value at the midpoint of the defined interval in each dimension and decay away from that position. And we set $\alpha = 1$ to achieve good performance based on our extensive testing.

$W$ determines the weight coefficients of the RBF neurons and the bias. With the purpose of enhancing the performance of the gradient flow and the convergence rate of neural network, the linear layer is initialized using the Xavier uniform distribution function provided by PyTorch library \cite{Glorot10}.

\subsection{Residual-based adaptive training dynamics}
Given a fixed set of training points $P$, a PIRBFNN with $N$ RBF hidden neurons can be trained to optimize the parameters $L$, $C$, and $W$ by minimizing the loss function using the L-BFGS optimizer with learning rate $\eta$.  

To further enhance the network's approximation capability, we propose a PDE residual-based adaptive refinement strategy that dynamically increases the number of RBF neurons during training. The key insight is to identify regions where the PDE residual is large and strategically place new RBF centers at those locations, thereby improving the network's ability to capture complex solution features while maintaining computational efficiency.

The adaptive strategy operates in two complementary phases: residual monitoring and network expansion. At each iteration, we randomly sample $s$ interior points from the solution domain and evaluate the PDE residuals. These residuals are aggregated into a decaying windowed average $\overline{\text{DWR}}_w^{(\tau)}$ over a window of size $w$ to track the training progress. To determine convergence, we evaluate a termination condition every $k$ iterations by analyzing the detrended residual change signal $\bm{D}$, which is obtained by removing the trend from the sequence of residual changes using a median filter. The algorithm terminates when both the positive and negative peak heights of $\bm{D}$ fall below a specified threshold $\epsilon$, indicating that the residual changes from adding neurons are not significant and the network has been sufficiently trained. Meanwhile, prior to satisfying the termination condition, the network is adaptively expanded at each of the $k$ iterations by adding $m$ new RBF neurons centered at the locations corresponding to the $m$ largest PDE residuals among the sampled points. This targeted placement ensures that computational resources are allocated to regions where the current approximation is most deficient.

In this adaptive learning scheme, when retraining the network after every $k$ iterations, the previously trained parameters are used as the initial values for the existing RBF neurons, thereby preserving the knowledge acquired during previous training iterations. While for the newly added RBF neurons, the initial centre locations are offered by the $m$ newly selected points with the largest PDE residuals. The weight coefficients and shape parameters are initialized according to the previously described initialization strategy. The detailed algorithmic procedure is presented in Algorithm~\ref{alg:adaptive_training}.


\begin{algorithm}[!h]
\caption{Residual-based adaptive training algorithm for PIRBFNN}
\label{alg:adaptive_training}
\begin{algorithmic}[1]
\Require Initial number of RBF neurons $N$, training points set $P = P_\mathcal{L} \bigcup P_T \bigcup P_\mathcal{B}$, sampling size $s$, refinement parameters $k$, $m$, window size $w$, convergence threshold $\epsilon$
\Ensure Trained PIRBFNN with optimized network structure
\State Initialize PIRBFNN with $N$ RBF neurons using the initialization strategy described in Section 3.4
\State Initialize L-BFGS optimizer with a learning rate $\eta$
\State Initialize iteration counter $\tau \leftarrow 0$
\While{not converged}
    \State $\tau \leftarrow \tau + 1$
    \State Train PIRBFNN on fixed training set $P$ using L-BFGS optimizer for one iteration
    \State Randomly sample $s$ interior points $\{(\bm S_j, t_j)\}_{j=1}^{s}$ from solution domain
    \State Evaluate PDE residuals: $r_j = |\mathcal{L} \hat{V}(\bm S_j, t_j) - f(\bm S_j, t_j)|$ for $j = 1, \ldots, s$
    \State Compute mean squared residual: $R^{(\tau)} = \frac{1}{s}\sum_{j=1}^{s} r_j^2$
    \State Compute decaying windowed residual average: $\overline{\text{DWR}}_w^{(\tau)} = \frac{1}{\tau}\sum_{i=\tau-w+1}^{\tau} R^{(i)}$,~~when 
    
    $\tau > w-1$ 
    \State Compute residual change: $\Delta R_w^{(\tau)} = \overline{\text{DWR}}_w^{(\tau)} - \overline{\text{DWR}}_w^{(\tau-1)}$,~~~~when $\tau > w-1$ 
    \If{$\tau \bmod k = 0$ \textbf{and} $\tau \neq 0$} 
        \State Extract window: $\bm{\Delta R}_w = [\Delta R_w^{(\tau-k+1)}, \ldots, \Delta R_w^{(\tau)}] \in \mathbb{R}^{k}$
        \State Compute trend using median filter: $\bm{T} = \text{MedianFilter}(\bm{\Delta R}_w,k) \in \mathbb{R}^{k}$
        \State Compute detrended signal: $\bm{D} = \bm{\Delta R}_w - \bm{T} \in \mathbb{R}^{k}$
        \State Find positive peak height: $h_+ = \max\{D_j : D_j > 0, j \in [1, k]\}$
        \State Find negative peak height: $h_- = \min\{D_j : D_j < 0, j \in [1, k]\}$
        \If{$h_+ \leq \epsilon$ \textbf{and} $|h_-| \leq \epsilon$}
            \State \textbf{break} \Comment{Stop condition satisfied: both peaks below threshold}
        \EndIf
    \EndIf
    \If{$\tau \bmod k = 0$ \textbf{and not converged}}
        \State Select $m$ points $\{(\bm S_{j^*}, t_{j^*})\}$ with largest residuals from the $s$ sampled points
        \State Add $m$ new RBF neurons with centres at these selected points and initialization strategy applied to the weights and shape parameters
        \State Retain trained parameters for existing $N$ neurons
        \State Update $N \leftarrow N + m$
        \State Reinitialize L-BFGS optimizer with updated network parameters
    \EndIf
\EndWhile
\State \Return Trained PIRBFNN
\end{algorithmic}
\end{algorithm}

\begin{figure}[!h]
	\centering
	\includegraphics[width=4.5in]{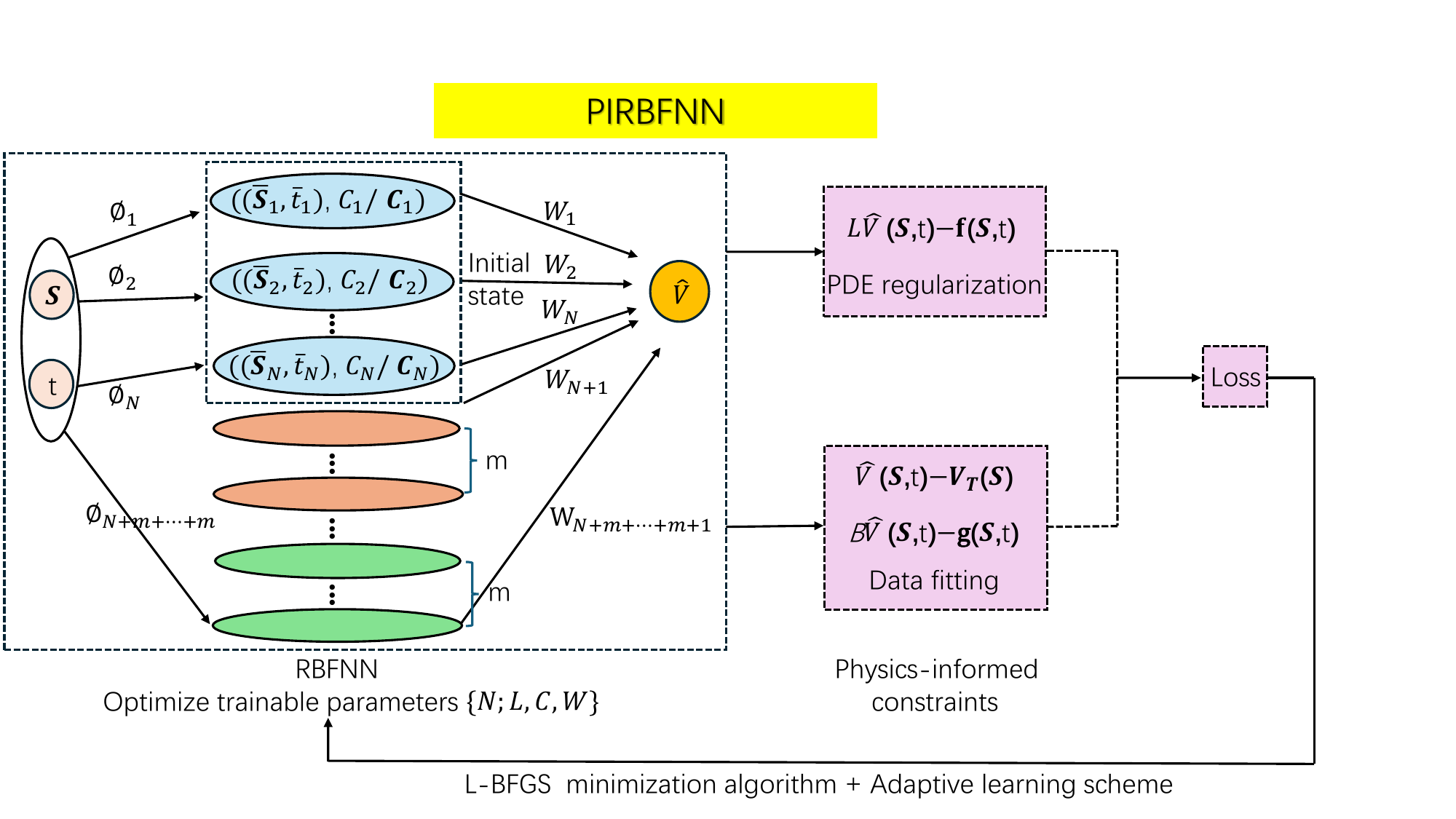}
	\caption{The architecture of an PIRBFNN for solving the TBVP \eqref{Gov_equation_general}--\eqref{Bou_condition_general}.}
	\label{fig:Architecture}
\end{figure}

A schematic diagram of an adaptively trained PIRBFNN for solving the TBVP \eqref{Gov_equation_general}--\eqref{Bou_condition_general} is shown in Fig.\ref{fig:Architecture}.

\subsection{Evaluation of the applicability of the network}
Upon inputting an arbitrary point from the definition domain into a trained network, the network generates the solution corresponding to that point. However, it has been demonstrated that a well-trained network with low loss value does not always guarantee the proximity of the approximate solution to the true solution \cite{Aditi2021}. To evaluate the spatial-temporal generalization capability of our proposed PIRBFNN beyond the training domain, we randomly sample $l$ points from the target region to serve as test points and monitor the training loss curve and the testing error curve simultaneously during the training process. These points are generated using the same strategy as the initial RBF centres. Furthermore, to assess the trained PIRBFNN network’s ability to generalize across different PDE model parameters, we also test its convergence behavior on a new financial parameter value, with a primary focus on the volatility of the underlying asset, using two approaches: training the network from scratch with the new parameter and fine-tuning a pretrained network corresponding to an old parameter.

\section{Numerical experiments}
In this section, we employ PIRBFNNs to address option pricing problems in one-, two- and four-dimensional settings under the Black–Scholes models, aiming to validate the feasibility and reliability of our proposed method. All experiments are executed in the VSCode for Python programming, on a single RTX 3090 24G GPU at a Linux server. To assess the accuracy of the approximation $\hat V$, the pointwise absolute error (PAE) and the root mean squared error (RMSE) are utilized as measurement metrics, which are defined as follows:

\begin{equation}
	PAE = \left|\hat V(\bm S_i,t_i) - V(\bm S_i,t_i)\right|, ~~(i=1,2,\cdots,l)
\end{equation}
\begin{equation}	
	RMSE = \sqrt{\frac{1}{l}\sum\limits_{i=1}^{l} \big(\hat V(\bm S_i,t_i)-V(\bm S_i,t_i)\big)^2}
\end{equation}
where $\hat V$ represents the PIRBFNN prediction, $V$ denotes either the exact solution or the reference data, and $l$ corresponds to the total number of test points.

\subsection{Example 1: Single-asset European put option}

The pricing of one-underlying asset European option within the Black-Scholes framework is a benchmark problem extensively studied in the finance community. In the case of the put option we consider, the governing Black-Scholes Eq.\eqref{BS_equation_general} is described by the one-dimensional PDE
\begin{equation}\label{1DB-S}
\frac{\partial V(S_1,t)}{\partial t} + \frac{1}{2} \sigma_1^2 S_1^2 \frac{\partial^2 V(S_1,t)}{\partial S_1^2} + r S_1 \frac{\partial V(S_1,t)}{\partial S_1} - r V(S_1,t) = 0,~~(S_1,t) \in [0,S_{max}] \times [0,T],
\end{equation}
equipped with the terminal payoff condition
\begin{equation}\label{1DPayoff}
V(S_1,T) = \max(K - S_1,0), 
\end{equation}
and the spatial boundary conditions
\begin{equation}\label{1Dboundary}
	\begin{cases} 
		V(0,t) = K e^{-r(T-t)},     \\ 
		V(S_{max},t) = 0.
	\end{cases} 
\end{equation}

The exact solution for the European put option is given by \cite{WOS:000898673800002}
\begin{equation}
	V_{exact}(S_1,t) = K e^{-r(T-t)} N(-d_2) - S_1 N(-d_1),
\end{equation}
where $N(\cdot)$ is the cumulative distribution function of a standard normal random variable $d$, defined as
\begin{equation*}
	N(d) = \frac{1}{\sqrt {2 \pi}} \int_{-\infty}^{d} e^{-\frac{1}{2} \xi^2} d \xi,
\end{equation*}
with $d_1 = \frac{log(\frac{S_1}{K}) + (r + \frac{1}{2} \sigma_1^2)(T-t)}{\sigma_1 \sqrt{T-t}},$
and $d_2 = d_1 - \sigma_1 \sqrt{T-t}.$

In the numerical implementation, we specify the model parameters as the following values: $\sigma_1 = 0.2, r = 0.05, \rho_{11} = 1, T = 0.5, K = 10$, and $S_{max} = 30$.

\subsubsection{PIRBFNN with a fixed number of neurons}
We employ $N = 1200$ RBF neurons in the hidden layer to design the PIRBFNN. For the learning process, we select $M = 2800$ training points. Among them, $M_\mathcal{L} = 400 * 4 = 1600$ points are located in the interior, where the factor $4$ refers to the number of boundaries of the two-dimensional spatial-temporal domain. In addition, $M_T = 400$ points are placed on the terminal time boundary, and $M_\mathcal{B} = 400 *2 = 800$ points are distributed along the spatial boundaries, where the factor $2$ corresponds to the two spatial boundary conditions of a single-asset European put option. The learning rate for the L-BFGS optimizer is set to $1$. A set of $l = 500$ interior points at time t = 0 is taken as the test points.

Remark: Given that equal weights are assigned to the PDE residual term and the boundary condition loss in the loss function, the number of boundary sampling points is chosen to be comparable to that of the interior sampling points here to prevent one physical constraint from dominating the optimization process and causing gradient imbalance. This empirically balances the influence of the two loss components on parameter updates, thereby enhancing the network’s capability to jointly approximate the solution within the domain and on the boundary, and improving the stability of the obtained solution. The same sampling strategy is also employed in the subsequent multi-dimensional examples.

Given that the seed value augmented in the random module determines both the generation of training set and the initialization of trainable parameters, we begin by statistically evaluating the impact of varying seed settings on the performance of PIRBFNN. Three infinitely smooth RBFs chosen as activation functions to construct the neural network are provided below: \\[5pt]
Gaussian function: $\phi (r) = e^{- C^2 r^2} = e^{- C^2 \lVert \bm X - \bm \bar X \lVert^2}$\\[5pt]
Inverse quadratic function: $\phi (r) = \frac{1}{1 + C^2 r^2} = \frac{1}{1 + C^2 \lVert \bm X - \bm \bar X \lVert^2}$\\[5pt]
Inverse multiquadric function: $\phi (r) = \frac{1}{\sqrt {1 + C^2 r^2}} = \frac{1}{\sqrt {1 + C^2 \lVert \bm X - \bm \bar X \lVert^2}}$\\[5pt]
where $\bm X$ and $\bm \bar X$ represent the collocation training point $(S_1,S_2,\dots,S_d,t)$ and centre point $(\bar S_1,\bar S_2,\dots,\bar S_d,\bar t)$, respectively. $C$ denotes the shape parameter information of the RBF neuron at the centre point. 

\begin{figure}[!ht]
    \centering
    \subfigure[]{
    \includegraphics[height=0.23\textwidth]{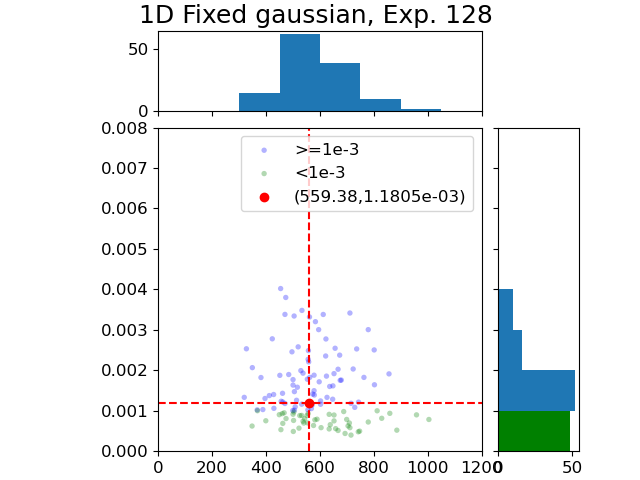}}%
    \subfigure[]{
    \includegraphics[height=0.23\textwidth]{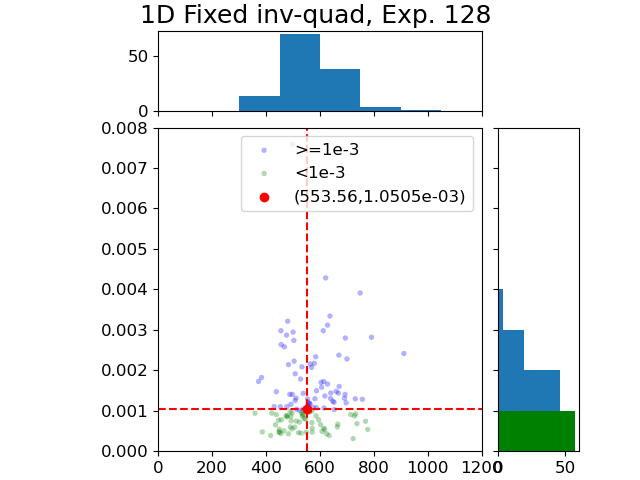}}%
    \subfigure[]{
    \includegraphics[height=0.23\textwidth]{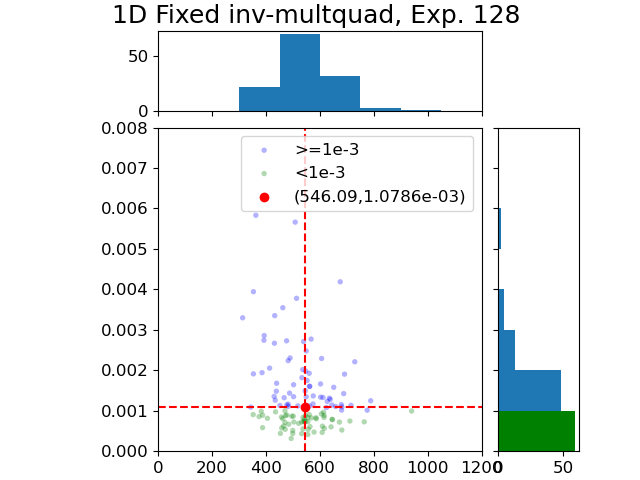}}%
\caption{Example 1: The results obtained from PIRBFNNs using different activation functions with respect to $128$ seed values. (a): Gaussian function, (b): Inverse quadratic function, (c): Inverse multiquadric function.}
    \label{fig:multiseedRBFneuron}
\end{figure}

Based on a sample set of $128$ distinct seeds, we accordingly train the PIRBFNNs and produce Fig.\ref{fig:multiseedRBFneuron} (a), (b), and (c) corresponding to the three different kernel functions mentioned above, where each point represents the network result obtained with a specific seed value. The horizontal and vertical coordinates of the point are the number of iterations and the final RMSE testing error at convergence of the network training (i.e., the error ceases to change), respectively. The top histogram displays the percentage of seed counts associated with each iteration numbers, while the right-side histogram illustrates the percentage of seed counts corresponding to various accuracy levels. Furthermore, a red point is plotted in each graph by first calculating the median, discarding the $25\%$ of samples with significant deviations from it, and then averaging the remaining values. This red point serves as an indicator of the network's average performance across all test samples. The results suggest that our PIRBFNN exhibits low sensitivity to seed variations. The accuracy remains stable, with errors consistently around the $10^{-3}$ to $10^{-4}$, and no notable deviations observed. Moreover, the large number of seeds offer the desired accuracy level of $10^{-4}$. Thereafter, to simplify the presentation, we would like to propose using a single seed to illustrate the subsequent findings in the following experiments.

\begin{figure}[!ht]
    \centering
    \subfigure[]{
    \includegraphics[height=0.17\textwidth]{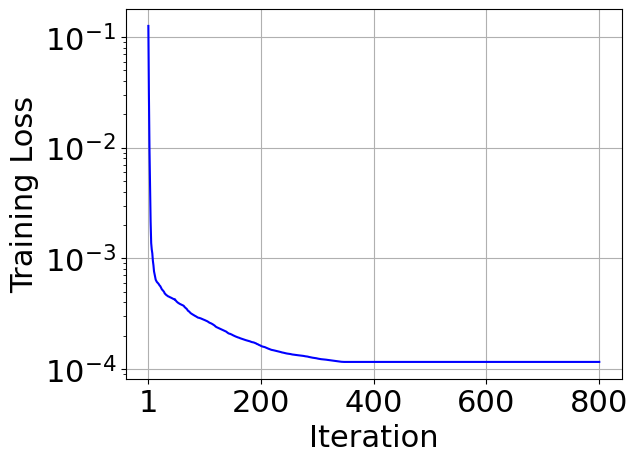}}%
    \subfigure[]{
    \includegraphics[height=0.17\textwidth]{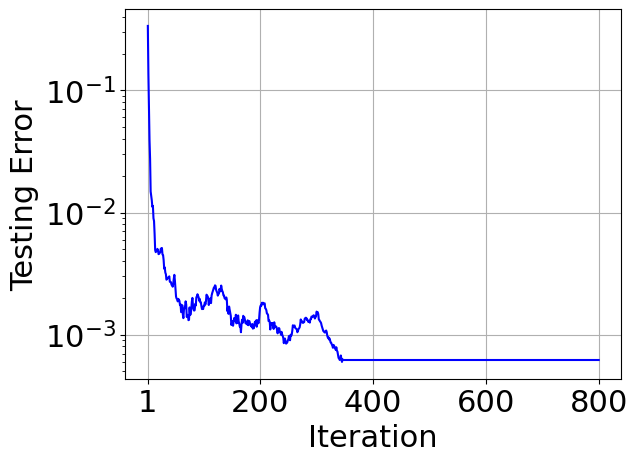}}%
    \subfigure[]{
    \includegraphics[height=0.17\textwidth]{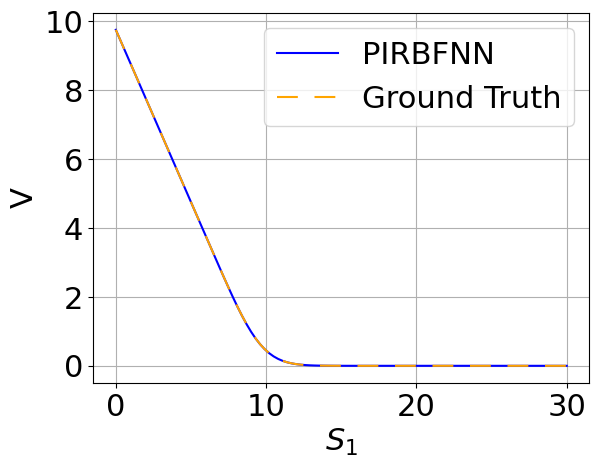}}%
    \subfigure[]{
    \includegraphics[height=0.17\textwidth]{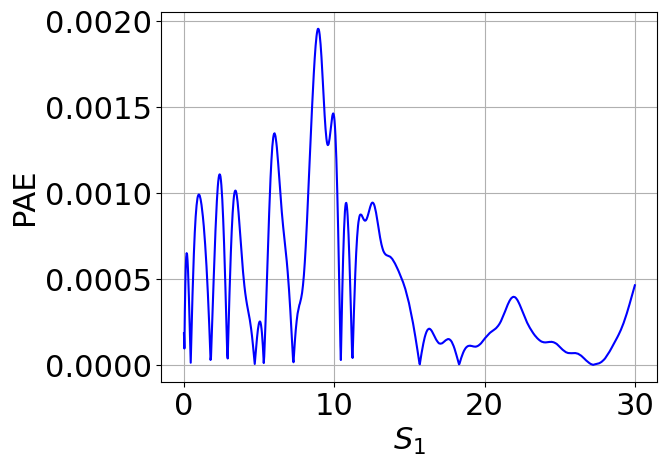}}
    \\
    \subfigure[]{
    \includegraphics[height=0.17\textwidth]{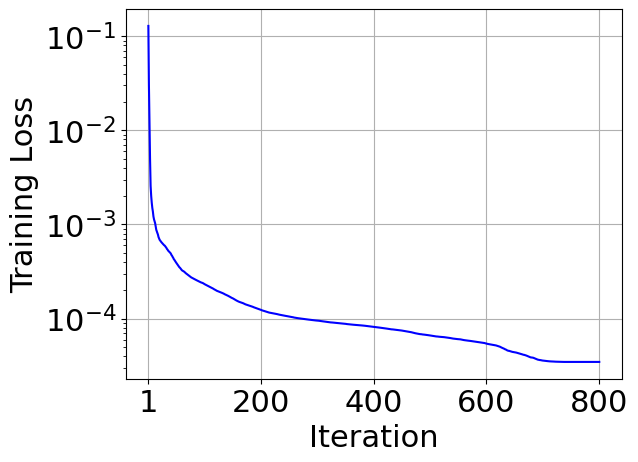}}%
    \subfigure[]{
    \includegraphics[height=0.17\textwidth]{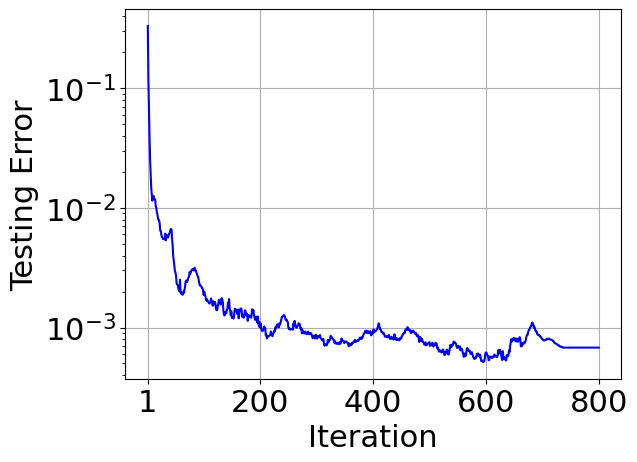}}%
    \subfigure[]{
    \includegraphics[height=0.17\textwidth]{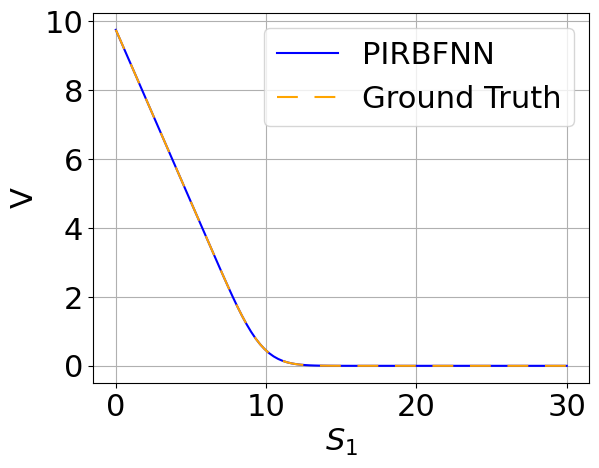}}%
    \subfigure[]{
    \includegraphics[height=0.17\textwidth]{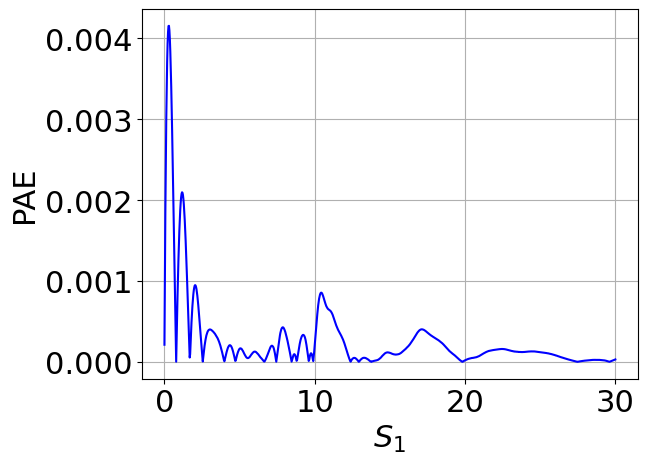}}
    \\
    \subfigure[]{
    \includegraphics[height=0.17\textwidth]{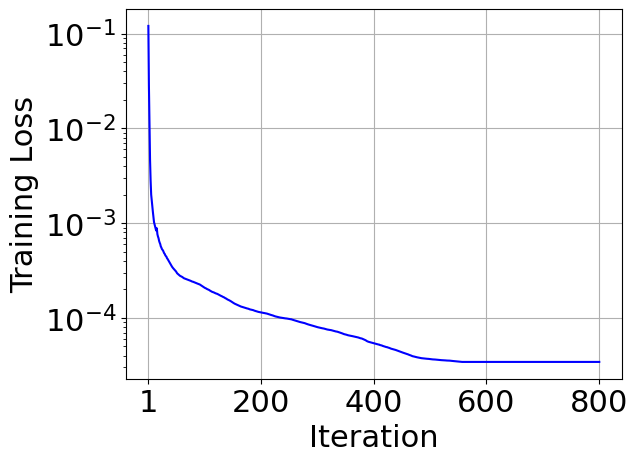}}%
    \subfigure[]{
    \includegraphics[height=0.17\textwidth]{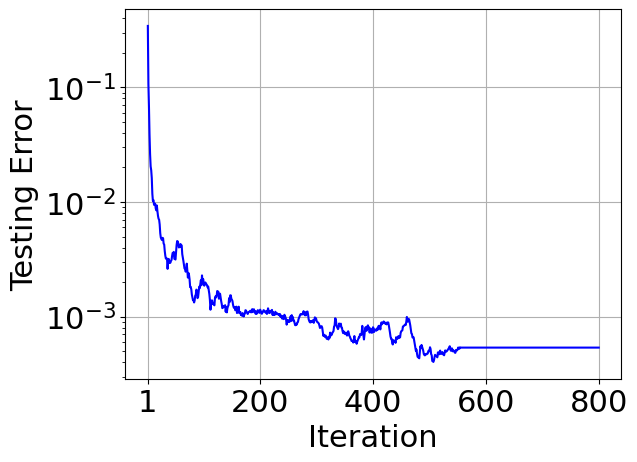}}%
    \subfigure[]{
    \includegraphics[height=0.17\textwidth]{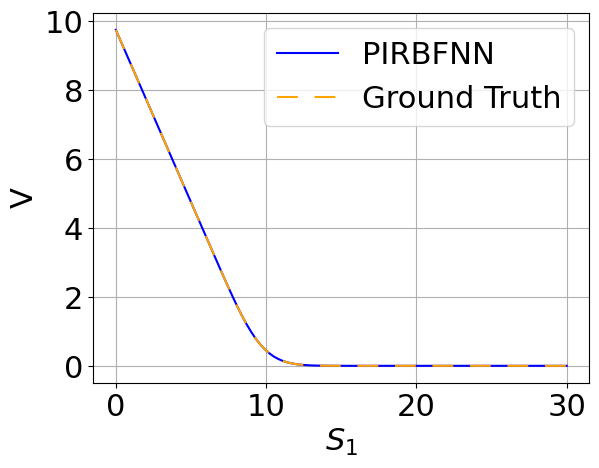}}%
    \subfigure[]{
    \includegraphics[height=0.17\textwidth]{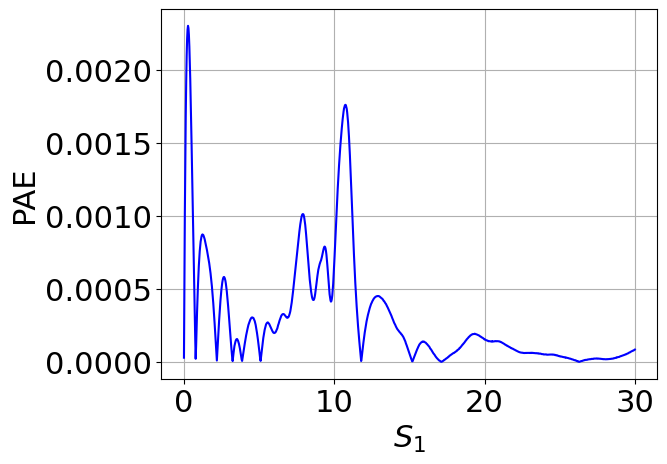}}
    \caption{Example 1: The results obtained from PIRBFNNs using different activation functions with respect to single seed value. Top panel: Gaussian function, middle panel: Inverse quadratic function, bottom panel: Inverse multiquadric function.}
    \label{fig:singleseedRBFneuron}
\end{figure}

Based on a given seed, the first row of Fig.\ref{fig:singleseedRBFneuron} presents the results obtained by using a PIRBFNN with the Gaussian function. In (a) and (b), the training loss value history and RMSE testing value history are illustrated with respect to the number of iterations during the learning process. In (c), a comparison is made between the approximate predictions after training and the exact solution. Additionally, (d) depicts the corresponding PAE curve. The second and last rows showcase the results of similar experiments, employing the inverse quadratic function and the inverse multiquadric function, respectively. As evident from the first two columns of the plot Fig.\ref{fig:singleseedRBFneuron}, we observe that the trend of the error value on the testing points is very similar to the one of the loss value on the training points, with a notable correlation: a lower training loss corresponds to a higher approximation accuracy. From the simulations presented in the third column, it is apparent that all these RBF activation functions can effectively solve this problem with good precision via the L-BFGS gradient decent optimizer. These experimental attempts underscore that PIRBFNN exhibits excellent representation accuracy and generalization ability. Furthermore, across all three types of RBF neurons considered, the PIRBFNNs consistently achieve high accuracy with stable convergence. For Gaussian function, at the $347$th iteration, the RMSE can reduce to $6.1967\times 10^{-4}$, with the largest PAE being just $1.9563\times 10^{-3}$. The inverse quadratic function and the inverse multiquadric function necessitate approximately $735$ iterations and $534$ iterations, respectively, to achieve training convergence and the same level of accuracy. Given the comparable performance of the three kernel functions, our subsequent work presents results only for the PIRBFNN employing the Gaussian activation function.

\subsubsection{PIRBFNN with adaptively added neurons}
\begin{figure}[!h]
	\centering
	\includegraphics[width=2.5in]{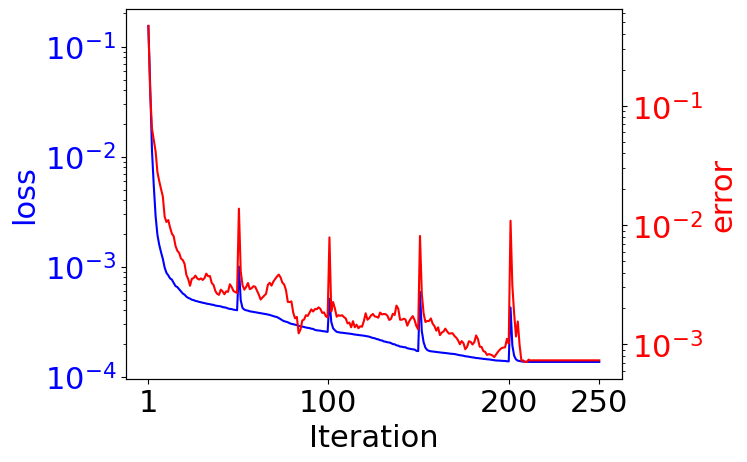}
    \caption{Example 1: Adaptive training of the PIRBFNN for solving single-asset European put option: loss value history (blue curve), RMSE value history (red curve).}
	\label{fig:Adaptive training}
\end{figure}

After that, to speed up the convergence rate of the PIRBFNN for effectively solving the option pricing problem \eqref{1DB-S}-\eqref{1Dboundary} without overfitting, we apply a PDE residual-based refinement scheme to dynamically increase the number of RBF hidden neurons during the learning process. While maintaining the learning rate unchanged, we first initialize the PIRBFNN with $N = 350$ RBF neurons and train the network on the previous training points. Subsequently, by utilizing the adaptive learning algorithm proposed in the preceding section, a total of $200$ additional RBF neurons are incrementally introduced to the hidden layer with $m = 50$ newly added neuron centres from $s = 1000$ randomly generated points within the solution domain at every $k = 50$th iteration. $w$ and $\epsilon$ are set to $48$ and $2\times10^{-6}$, respectively, as the checking parameters for iteration termination. Based on the same testing points utilized in the above experiment, Fig. \ref{fig:Adaptive training} depicts the variations in the training loss values and testing error values over iterations during the adaptive learning process, represented by the blue and red curves, respectively. Despite the spiky fluctuations in the training process caused by the initialization of newly introduced trainable parameters when increasing the number of neuron centres, the network quickly stabilized as the iterative optimization progressed. Compared to Fig. \ref{fig:singleseedRBFneuron} (a) and (b), it is evident that with a final network size of only $550$ RBF neurons, the same level of accuracy in terms of RMSE can be achieved at the $250th$ iteration when the training terminates. Without compromising accuracy, the number of iterations required for convergence as well as the overall training time have been effectively reduced. This observation validates that simultaneously fine-tuning the network parameters while optimizing the network structure can enhance the computational accuracy and training efficiency of our PIRBFNN.

\begin{figure}[!h]
\centering  
\subfigure[]{
    \includegraphics[height=1in]{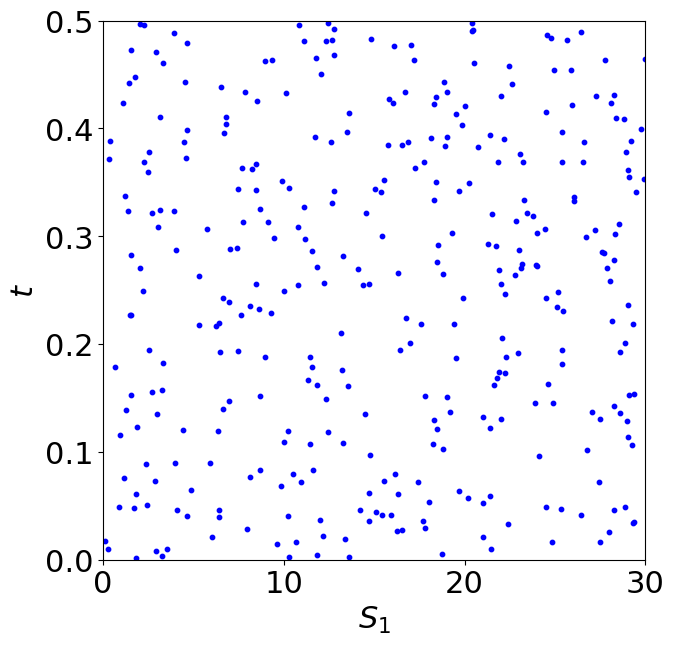}}%
\subfigure[]{
    \includegraphics[height=1in]{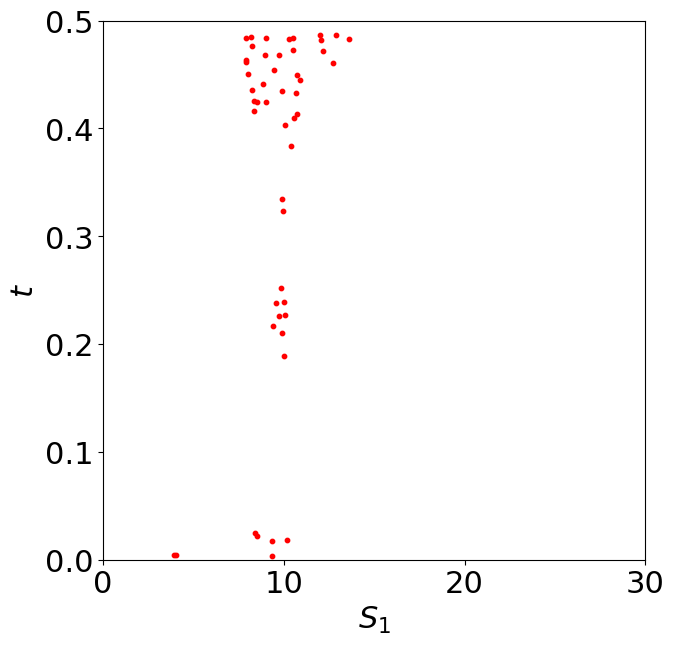}}%
\subfigure[]{
    \includegraphics[height=1in]{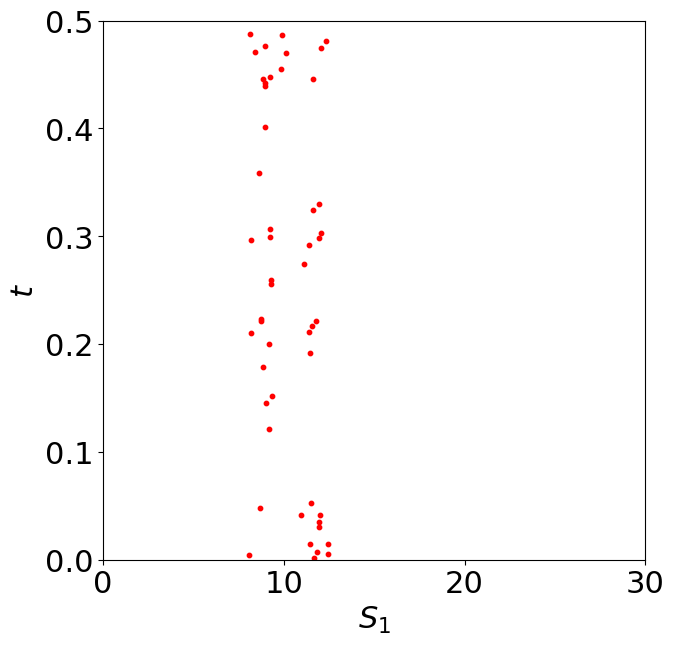}}%
\subfigure[]{
    \includegraphics[height=1in]{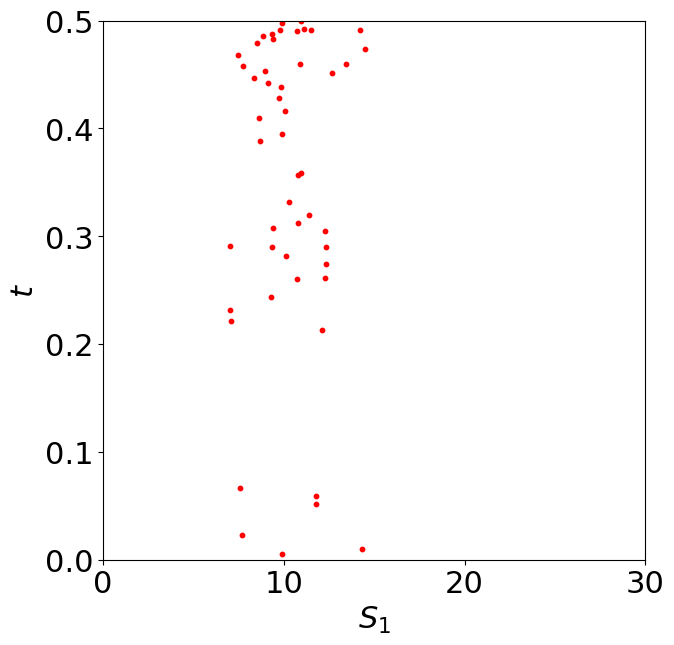}}%
\subfigure[]{
    \includegraphics[height=1in]{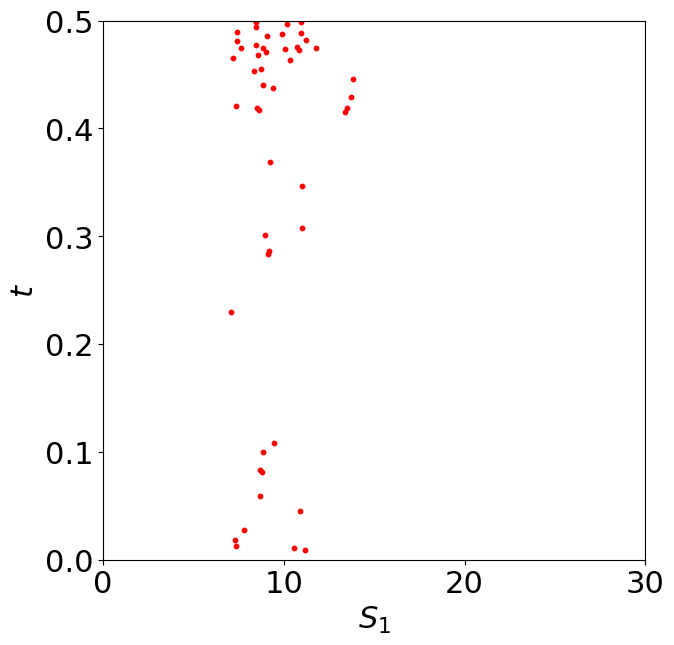}}%

\subfigure[]{
    \includegraphics[height=1in]{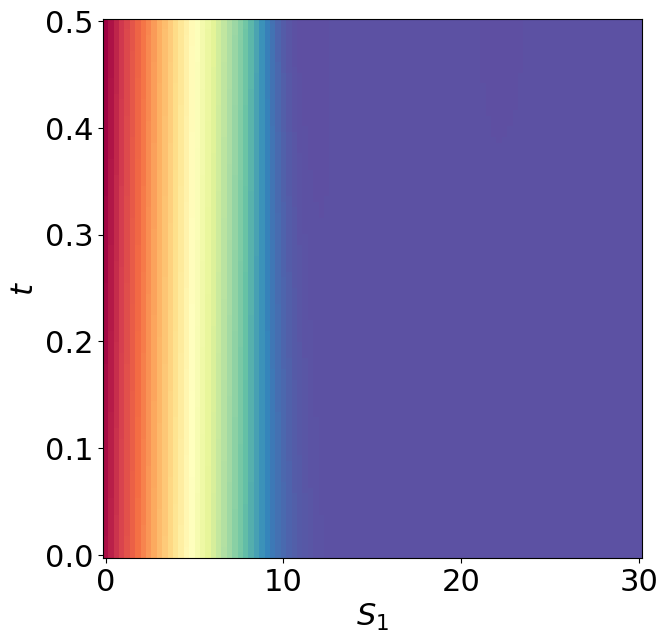}}%
\subfigure[]{
    \includegraphics[height=1in]{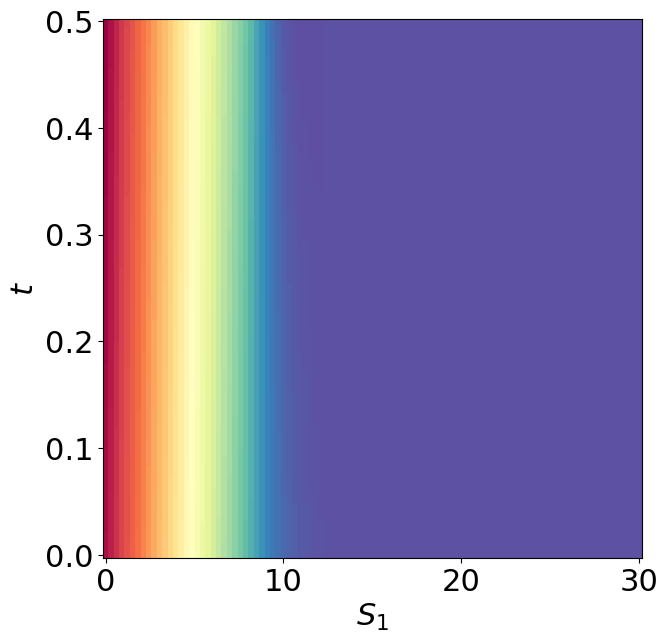}}%
\subfigure[]{
    \includegraphics[height=1in]{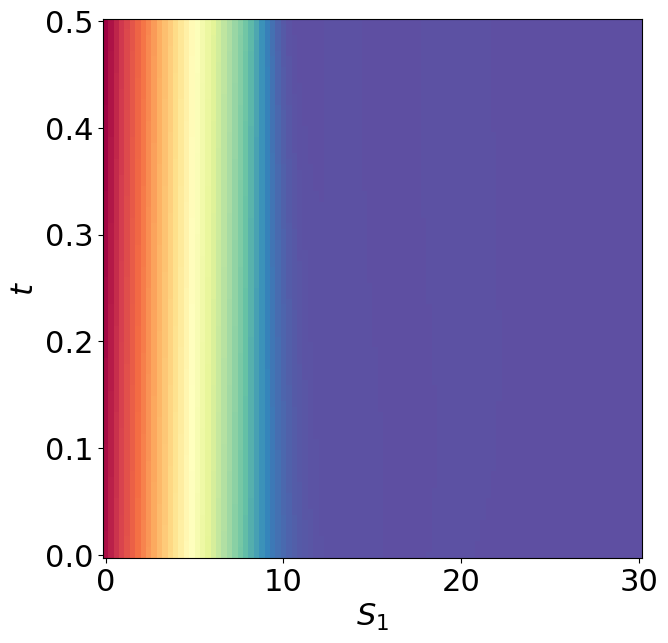}}%
\subfigure[]{
    \includegraphics[height=1in]{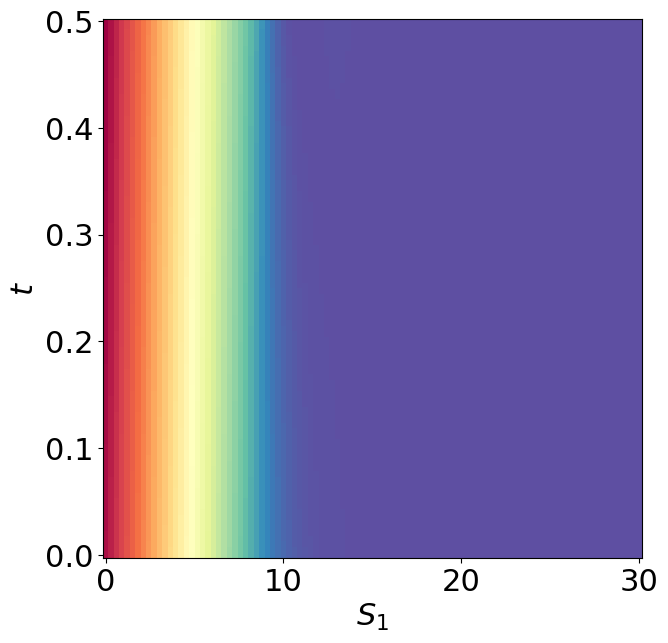}}%
\subfigure[]{
    \includegraphics[height=1in]{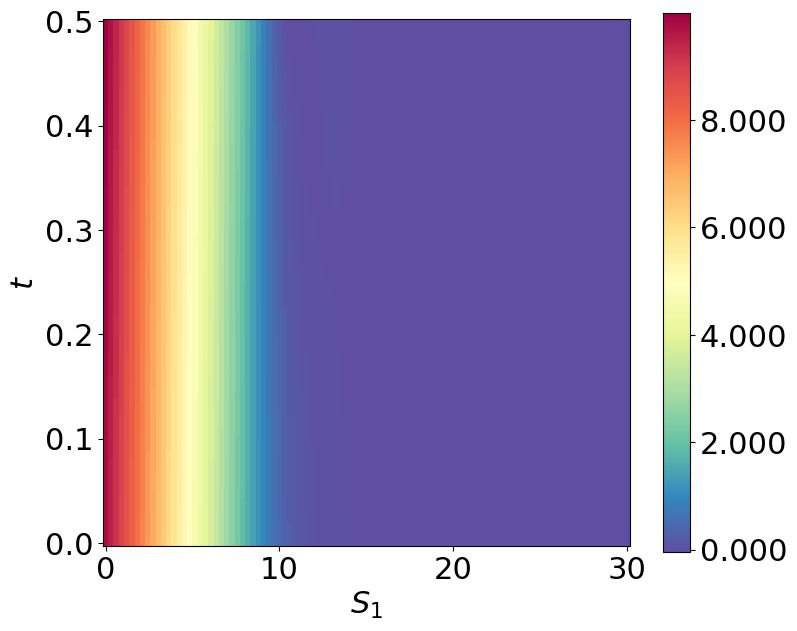}}%

\subfigure[]{
    \includegraphics[height=1in]{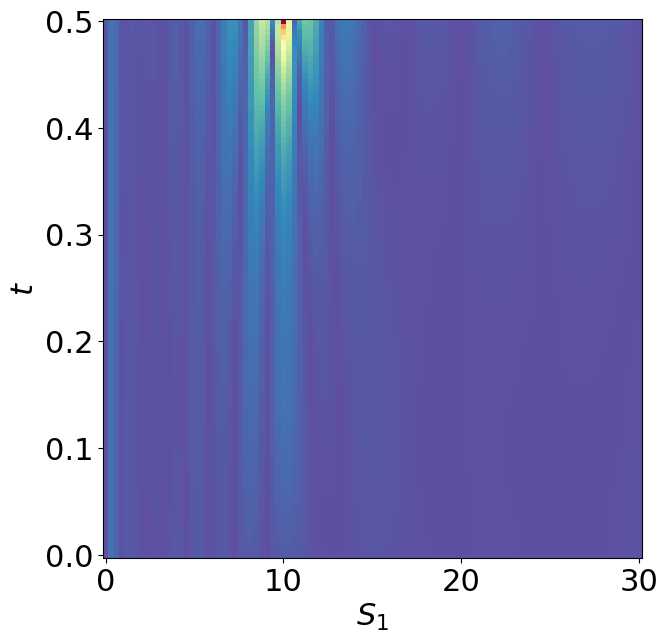}}%
\subfigure[]{
    \includegraphics[height=1in]{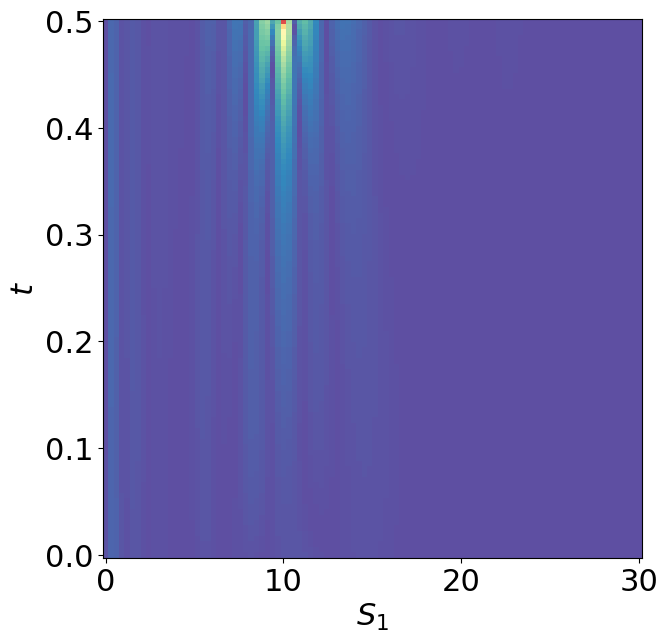}}%
\subfigure[]{
    \includegraphics[height=1in]{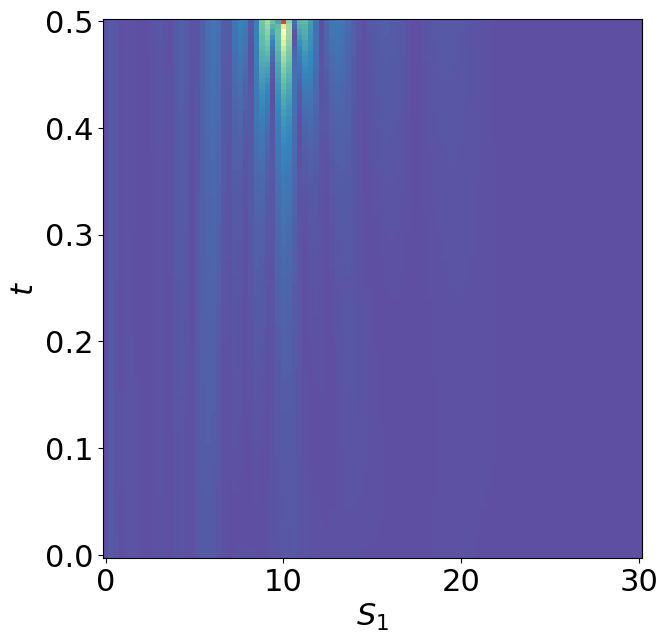}}%
\subfigure[]{
    \includegraphics[height=1in]{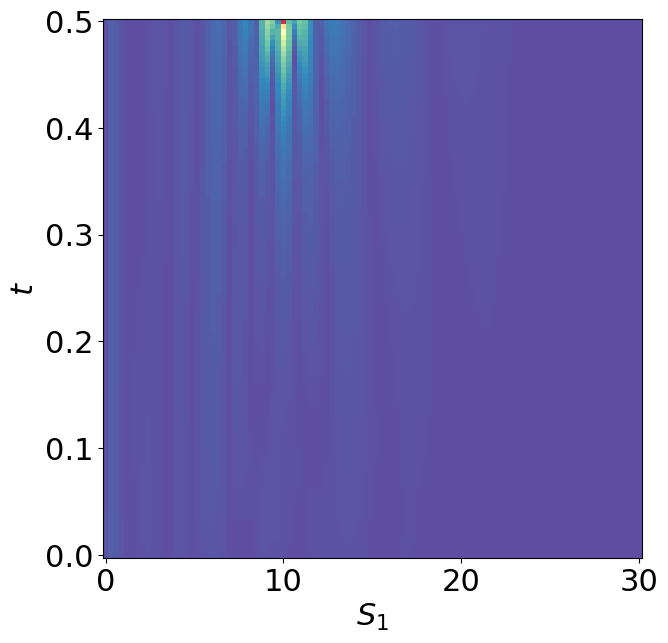}}%
\subfigure[]{
    \includegraphics[height=1in]{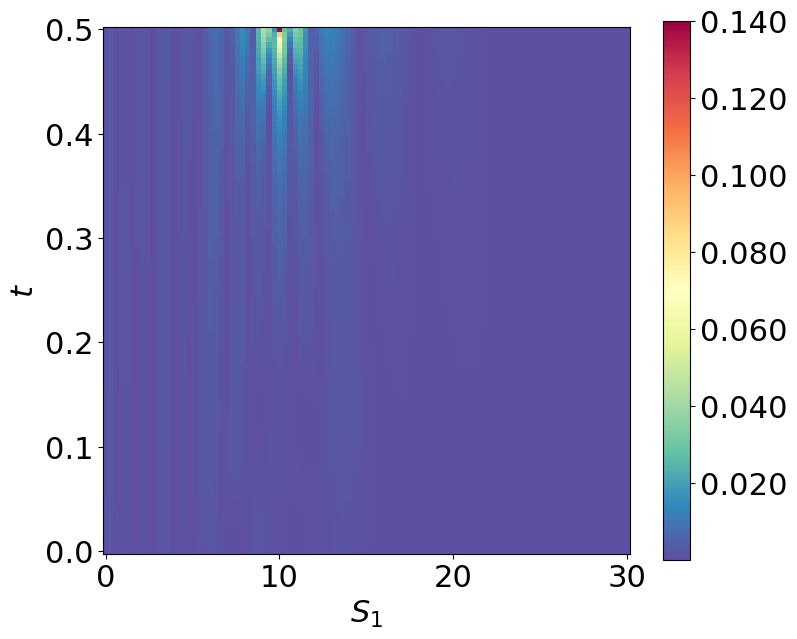}}%

\caption{Example 1: The results obtained from PIRBFNNs using varying numbers of RBF hidden neurons during the adaptive learning process. Column 1: initial $350$ neurons, columns 2-4: increase by $50$ neurons every $50$ iterations.}
	\label{fig:AdpRBFneuron}
\end{figure} 

Fig.\ref{fig:AdpRBFneuron} demonstrates the outcomes achieved by the PIRBFNNs employing various numbers of RBF hidden neurons during the above adaptive learning process, including the spatial distribution of centre points for the initial and newly added RBF neurons, pointwise option price predictions across the entire spatial-temporal domain, and the corresponding PAE values. The first column presents the results from the network trained with an initial N = 350 neurons at the k = 50th iteration. The second to fifth columns show the results after training the network with 400, 450, 500, and 550 neurons at the 100th, 150th, 200th, and 250th iteration. The learning finished at the 250th iteration. Given the non-smooth terminal condition at the strike price, the option price exhibits low regularity in the spatial-temporal region near $K = 10$ with time approaching $T = 0.5$, resulting in the largest error. But when incorporating the PDE residual information to dynamically introduce new neurons for architecture optimization of PIRBFNN, we observe that the greater the number of hidden neurons in the network, the higher the level of accuracy in the approximation. Nevertheless, we would like to clarify here that, for the given amount of training points, excessive RBF neurons do not invariably improve model performance furthermore. On the contrary, overfitting may occur, resulting in a reduction of accuracy and an increase of computation. However, our proposed learning strategy adaptively increases the number of hidden neurons, allowing the network to expand to an appropriate width while ensuring high-accuracy output upon training termination. Moreover, the first row reveals that the PIRBFNN with adaptive training dynamics can capture the optimizer’s attention to the challenging region quite well. More neurons are added near the problematic place to decrease the errors consistently as expected, which indicates the effectiveness of our learning scheme.

\begin{figure}[!h]
	\centering
	\includegraphics[width=2.0in]{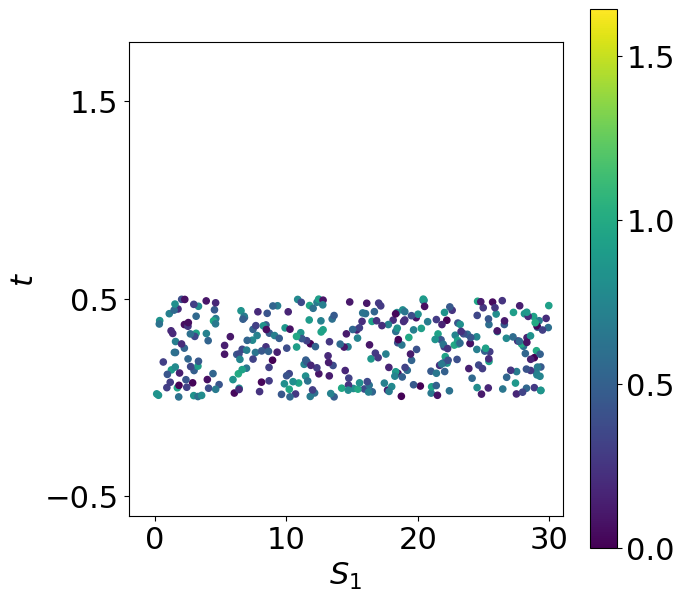}
	\includegraphics[width=2.0in]{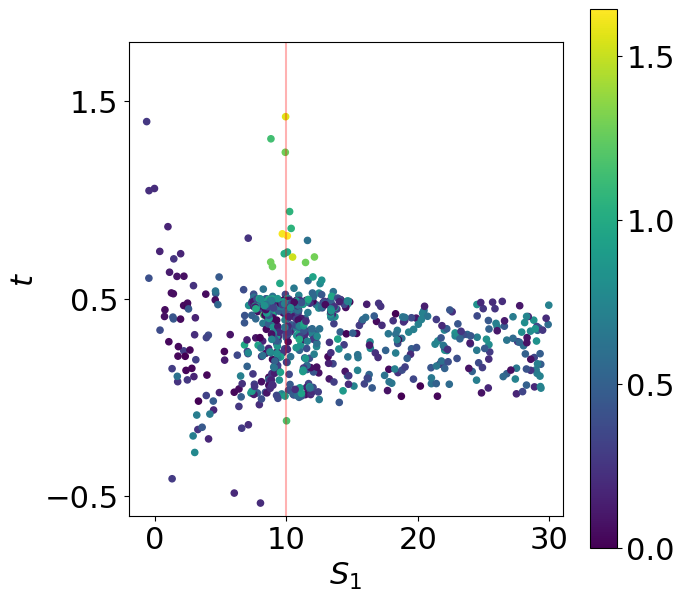}
	\caption{Example 1: Parameters of RBF neurons (a)initial state before training, (b) updated state after training.}
	\label{fig:Training}
\end{figure}

\begin{figure}[!h]
	\centering
    \subfigure[]{
    \includegraphics[width=2.0in]{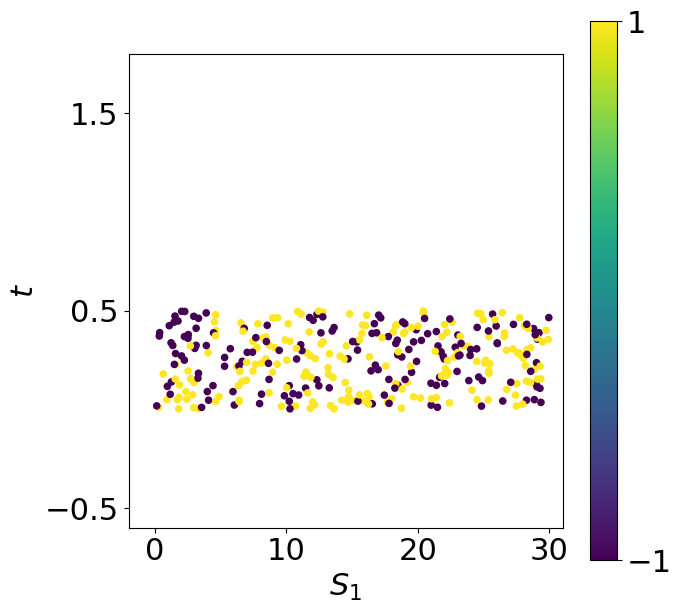}
    }
	\subfigure[]{
    \includegraphics[width=2.0in]{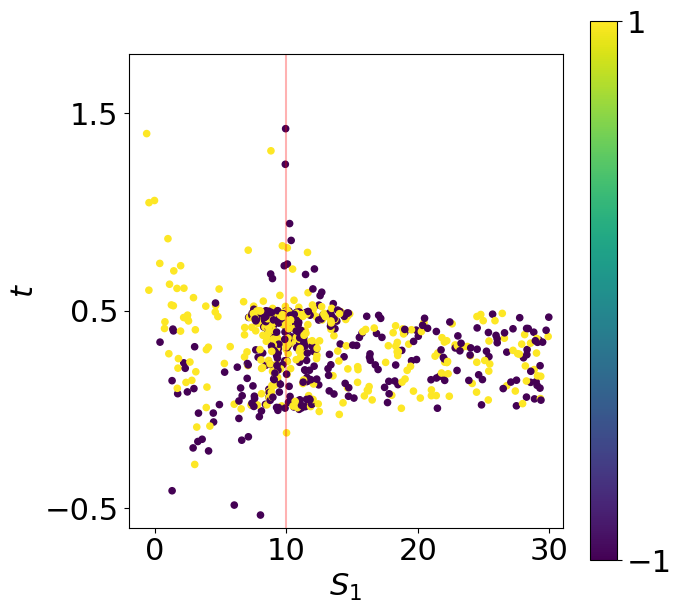}
    }
    \qquad
    \subfigure[]{
    \includegraphics[width=2.0in]{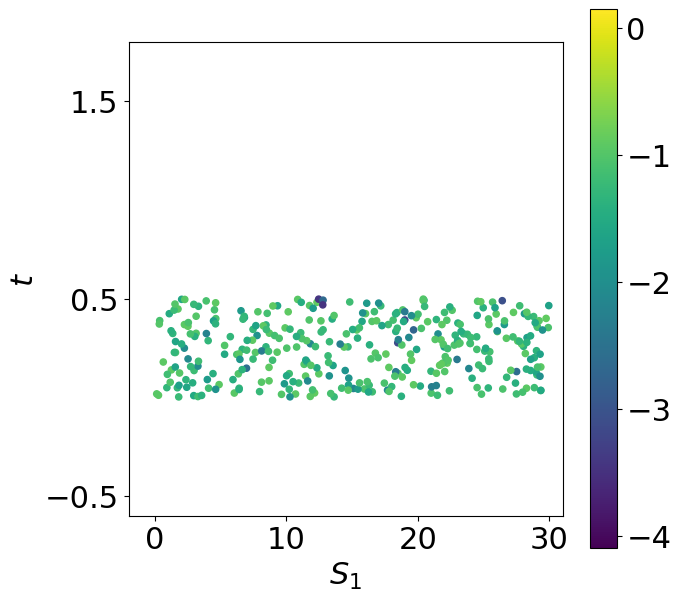}
    }
	\subfigure[]{
    \includegraphics[width=2.0in]{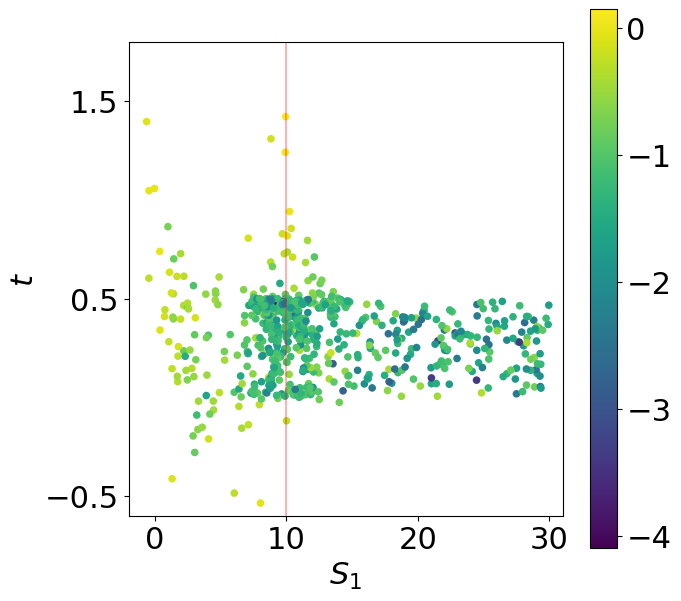}
    }
	\caption{Example 1: Linear weights of RBF neurons: initial state before training for (a) sign(W) and (c) $ \log_{10} \lvert W \rvert$, updated state after training for (b) sign(W) and (d) $ \log_{10} \lvert W \rvert$.}
	\label{fig:Training-2}
\end{figure}

Fig.\ref{fig:Training} (a) and (b) visualize the preliminary and final parameters of RBF neurons studied in the Fig.\ref{fig:AdpRBFneuron}, respectively, where each circle represents an RBF neuron, with circle number representing the neuron quantity, circle position displaying the centre location, and color intensity reflecting the absolute-valued shape parameter. Fig.\ref{fig:Training-2} also illustrates the states of the linear weights associated with each RBF neuron before and after training, highlighting both their signs and magnitudes. This experiment demonstrates the effectiveness of machine learning dynamics, which enables the adjustment of trainable parameters to optimal settings, yielding good approximation. In particular, regions characterized by low regularity condition exhibit a more pronounced response to the training effect. Furthermore, an intriguing observation is also noticed that the centres distributed beyond the original solution domain after training. This phenomenon aligns with the finding of the ghost point method proposed in \cite{WOS:000499761700018}, which verified that extending the centres to a region covering the computational domain can significantly enhance the accuracy of the traditional RBF collocation method.
  
\begin{figure}[!h]
	\centering
    \subfigure[]{
	\includegraphics[width=2in]{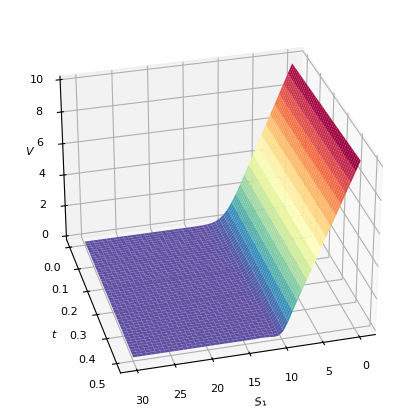}}
    \subfigure[]{
	\includegraphics[width=2.5in]{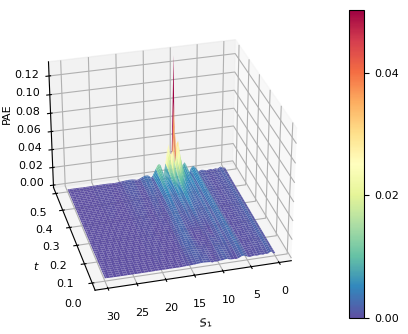}}
	\caption{Example 1: The time evolution profiles of single-asset European put option for (a) approximate prices, (b) PAE values.} 
	\label{fig:EvolutionProfile}
\end{figure}

Based on the previous numerical analysis, we can infer that the RBFNN trained by the adaptive learning dynamics using physics-informed loss function can effectively address single-asset European put option pricing problem. From the time evolution graphs of predicted option prices and corresponding PAE values after training shown in Fig.\ref{fig:EvolutionProfile} (a) and (b), respectively, it is clear to see that the proposed method has successfully mitigated the oscillation arising from the non-smoothness of the payoff function at the strike price as the terminal time is approached, yielding a smooth solution profile.

\begin{figure}[!h]
	\centering
    \subfigure[]{
	\includegraphics[width=2.5in]{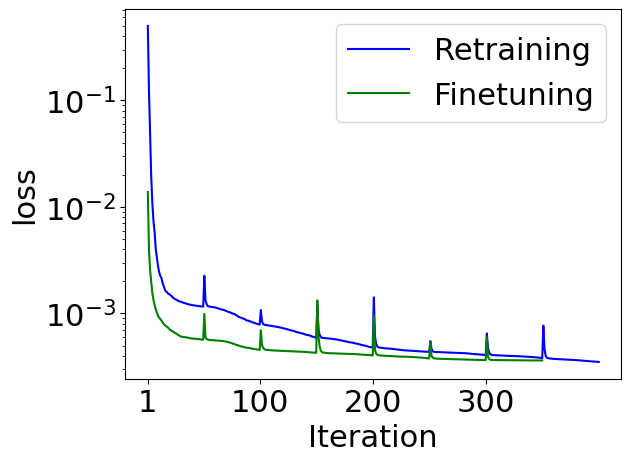}}
    \subfigure[]{
	\includegraphics[width=2.5in]{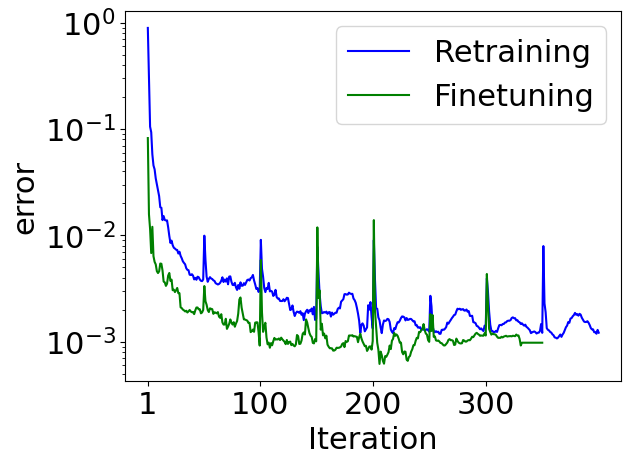}}
	\caption{Example 1: Illustrations of (a) training loss curves and (b) testing error curves for two experiments related to volatility.} 
	\label{fig:sigma}
\end{figure}

Given the practical applicability of option pricing problem in real-world scenarios, we conduct additional evaluations of our proposed neural network learning system under varying financial parameters that capture diverse market conditions. Here we mainly demonstrate numerical performance by focusing on the volatility of the underlying asset to conclude the discussion of this part. While keeping all other settings of the previous adaptive algorithm unchanged, we only adjust the value of $\sigma_1$ from $0.2$ to $0.3$. In the first experiment (illustrated as blue curves), the PIRBFNN is trained from scratch under $\sigma_1 = 0.3$ until the adaptive learning process terminates. In the second experiment (shown as green curves), we initialize the PIRBFNN using the architecture and trained parameters from the fully converged adaptive learning system at $\sigma_1 = 0.2$, then update only $\sigma_1$ to $0.3$ and resume adaptive training until convergence. Fig.\ref{fig:sigma} (a) shows the training loss curves versus optimization iterations for both experiments, while Fig.\ref{fig:sigma} (b) presents the corresponding testing error curves across the same iterations. Following a change in the PDE model parameter related to physical information, the RBFNN trained from scratch using the adaptive strategy converges to a computational accuracy of $1.2168 \times 10^{-3}$ after $400$ iterations. In comparison, resuming adaptive training from the previously converged RBFNN under the old parameter setting reaches an accuracy of $9.7956 \times 10^{-4}$ in only $350$ iterations. The results show that fine tuning the network that is continuing from the $\sigma_1 = 0.2$ model with $\sigma_1$ switched to $0.3$, achieves both a higher level of predictive performance and faster convergence than retraining from scratch. This provides strong evidence of the broad applicability of our network architecture in practical market contexts. 

\subsubsection{Comparative analysis of PIRBFNN with PINN}
\begin{figure}[!h]
	\centering
    \subfigure[]{
	\includegraphics[width=2.5in]{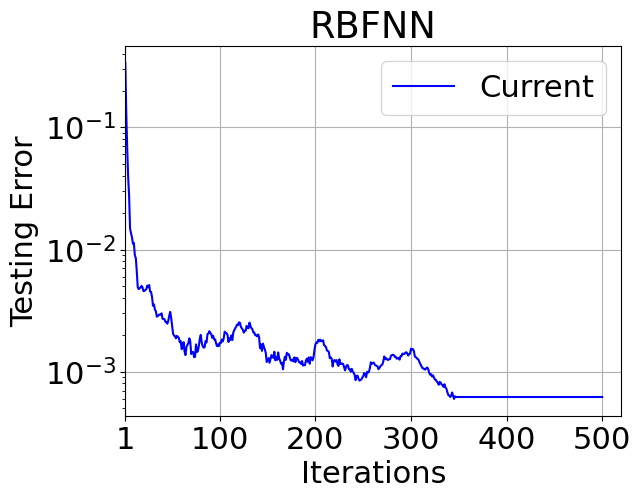}}
    \subfigure[]{
	\includegraphics[width=2.5in]{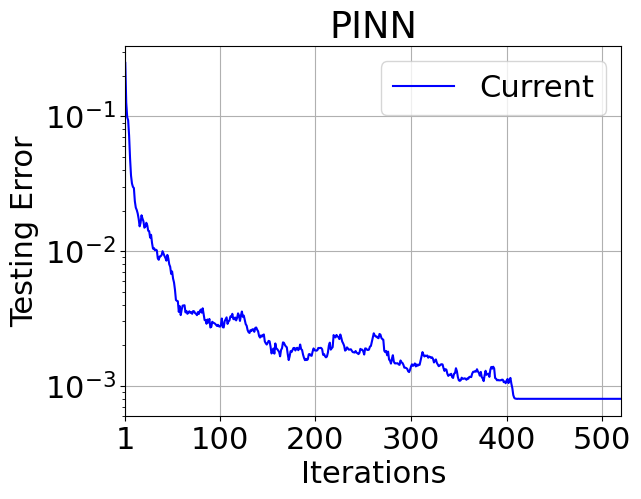}}
	\caption{Example 1: Comparison between PIRBFNN and PINN: RMSE versus the number of training iterations.}
	\label{fig:Comparison}
\end{figure}

Although the PIRBFNN possesses only a single hidden layer lacking deep learning property, the typical RBF neuron with centre distribution, shape controller, radial nonlinear mapping capability still conveys a wealth of information, thereby enhancing the representative ability of the neural network. Concerning the supportive evidence about this argument, we evaluate the performances of a PIRBFNN with Gaussian activation function and a PINN with tanh activation function with respect to the number of training iterations. Based on the $\sigma_1 = 0.2$ model and same training set as the previous experiments, the PIRBFNN is constructed with $1200$ hidden neurons to define its network architecture. In each iteration, a total of $4 \times 1200+1$ parameters need to be trained. While PINN is designed accordingly with a comparable number of trainable parameters as PIRBFNN, which informs the configuration of $2$ hidden layers and $67$ neurons in each. Fig.\ref {fig:Comparison} illustrates the trend of RMSE between the network prediction and the exact solution at $l = 500$ test points when $t = 0$ with respect to the number of training iterations, where (a) and (b) represent the outputs of the PIRBFNN and PINN, respectively. Upon reaching convergence of the training process, the PIRBFNN achieves an RMSE of $6.1967 \times 10^{-4}$ using $347$ iterations, whereas the PINN requires more than $400$ iterations to attain the RMSE in the level of $7.9952 \times 10^{-4}$. Compared to the PINN, PIRBFNN achieves faster convergence rate and higher computational accuracy.

\subsection{Double-asset exchange option}
An exchange option is a simple type of two-underlying assets option, enabling the holder to swap one financial asset for another upon expiration. Due to the absence of a strike price term, the conventional categorization of call and put options is not applicable to this particular case. When we consider the first asset $S_1$ as being long a call option and the second asset $S_2$ as being short a put option, the exchange option pricing problem can be modeled by the governing Black-Scholes Eq.\eqref{BS_equation_general} in the following form
\begin{equation}\label{2DB-S}
	\begin{array}{l}
	\frac{\partial V(S_1,S_2,t)}{\partial t} + \frac{1}{2} \sigma_1^2 S_1^2 \frac{\partial^2 V(S_1,S_2,t)}{\partial S_1^2} + \frac{1}{2} \sigma_2^2 S_2^2 \frac{\partial^2 V(S_1,S_2,t)}{\partial S_2^2} - \rho_{12} \sigma_1 \sigma_2 S_1 S_2 \frac{\partial^2 V(S_1,S_2,t)}{\partial S_1 \partial S_2} + r S_1 \frac{\partial V(S_1,S_2,t)}{\partial S_1}\\
	+ r S_2 \frac{\partial V(S_1,S_2,t)}{\partial S_2} - r V(S_1,S_2,t) = 0,~~~~~~~~~~(S_1,S_2,t) \in [0,S_{max}]^2 \times [0,T],
	\end{array}
\end{equation}
equipped with the terminal payoff condition
\begin{equation}\label{2Dpayoff}
		V(S_1,S_2,T) = \max(S_1 - S_2,0),
\end{equation}
the zero-side boundary conditions
\begin{equation}\label{2Dzeroboundary}
	\begin{cases} 
		V(0,S_2,t) = 0,  \\ 
		V(S_1,0,t) = S_1,
	\end{cases} 
\end{equation}
and the far-side boundary conditions
\begin{equation}\label{2Dfarboundary}
	\begin{cases} 
		V(S_{max},S_2,t) = V_{exact}(S_{max},S_2,t),  \\ 
		V(S_1,S_{max},t) = V_{exact}(S_1,S_{max},t),  
	\end{cases} 
\end{equation}
where $V_{exact}(S_1,S_2,t)$ represents the exact solution of the exchange option price at time $t$, given the asset prices $S_1$ and $S_2$. In the far-field boundary conditions, the exact solution is employed instead of the payoff function to obtain more accurate results for numerical testing. As presented in \cite{WOS:000898673800002}, the following analytical formula describes this relationship: 
\begin{equation}
	V_{exact}(S_1,S_2,t) = S_1 N(d_1) - S_2 N(d_2)
\end{equation}
where $N(\cdot)$ is the cumulative distribution function of the standard normal distribution with\\
$d_1 = \frac{log(\frac{S_1}{S_2}) + (\frac{1}{2} \sigma^2)(T-t)}{\sigma \sqrt{T-t}}, d_2 = d_1 - \sigma \sqrt{T-t}$, and $\sigma = \sqrt {\sigma_1^2 + \sigma_2^2 - 2 \rho_{12} \sigma_1 \sigma_2}$. 

For the numerical calculation, we employ the following model parameters: $\sigma_1 = \sigma_2 = 0.2, r = 0.05, \rho_{11} = \rho_{22}=1, \rho_{12} = \rho_{21} = 0.5, T = 1$, and $S_{max} = 40$.

We utilize an initial configuration of N = 2000 Gaussian RBF neurons in the hidden layer to construct the PIRBFNN. For the learning process, we choose $M = 8800$ collocation training points. These consist of $M_\mathcal{L} = 800*6 = 4800$ interior points, where the factor $6$ denotes the number of boundaries of the three-dimensional spatial-temporal domain; $M_T = 800$ points positioned on the terminal time boundary; and $M_\mathcal{B} = 800 * 4 = 3200$ points sampled on the spatial boundary planes, where the factor $4$ reflects the four spatial boundary conditions of a double-asset exchange option.  Additionally, we specify the learning rate of $1$ for the L-BFGS optimizer.

Following the proposed adaptive learning scheme, we commence learning of the initial PIRBFNN on the given training set. Subsequently, we monitor the $\overline{\text{DWR}}_w^{(\tau)}$ value at each optimization iteration $\tau$ and accordingly conclude the training process with the checking parameters $\epsilon \ = 2\times 10^{-4}$ and $w = 128$. Concurrently, the network’s width is expanded at every $k = 200th$ iteration by selecting $m = 200$ new points exhibiting the highest PDE residuals from a set of $s = 1000$ randomly sampled points within the solution domain. These points serve as the centre points of the newly inserted RBF hidden neurons. This adaptively adding neurons procedure is repeated every $k$ iterations until the training stopping criterion is satisfied. A set of $l = 500$ interior points at the time level $t = 0$ is chosen as the test points. 

\begin{figure}[!h]
	\centering
    \subfigure[]{
	\includegraphics[width=2.5in]{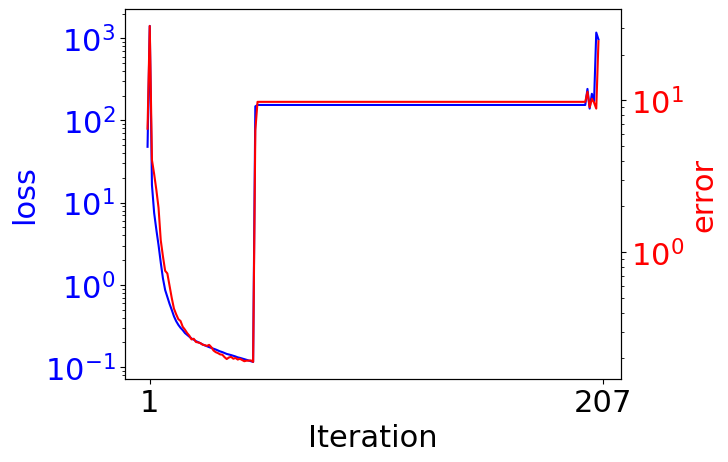}}
    \subfigure[]{
	\includegraphics[width=2.5in]{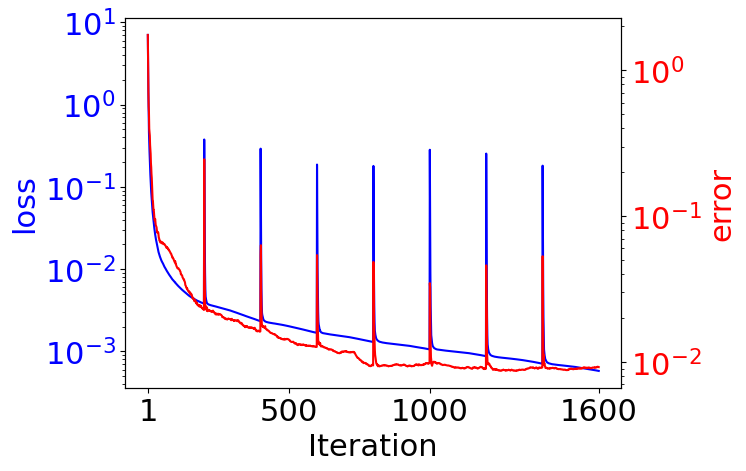}}
	\caption{Example 2: Adaptive training of the PIRBFNN employing (a) scalar shape parameters and (b) vectorized shape parameters for pricing double-asset exchange option: loss value history (blue curve), RMSE value history (red curve).} 
	\label{fig:Adaptive training2}
\end{figure}

Using the exact solution of the exchange option as a benchmark for RMSE evaluation,  Fig.\ref{fig:Adaptive training2} $(a)$ illustrates the histories of loss values and error values in relation to the number of iterations throughout the adaptive learning process, employing the same scalar shape parameters as in the single-asset case. In contrast, Fig.\ref{fig:Adaptive training2} $(b)$ depicts the corresponding experiment conducted with the shape parameters in vector form. From the figures, it is evident that the scalar shape parameters perform poorly in simulating two-dimensional exchange option pricing, exhibiting low predictive accuracy and divergent behavior. However, the trained PIRBFNN equipped with the vectorized shape parameters achieves exceptional generalization capability on the testing points. As the training loss curve decreases steadily, there is a consistent decline in the error curve. Upon termination of the adaptive training process, the RMSE has been reduced to $9.2078\times 10^{-3}$. Therefore, in this manuscript, the vectorized shape parameters are employed in multi-dimensional option models to enhance the representational capacity of the PIRBFNN, and all subsequent experiments are carried out based on this network architecture.

During this process, we also display the numerical results for various $S_1$ and $S_2$ values at time $t = 0$ predicted by the PIRBFNNs with different numbers of RBF hidden neurons in Table \ref{tab:Exchoptionvalue}. This highlights the impact of the number of hidden neurons on network performance. A progressive improvement in accuracy is seen with the increased number of neurons, up to a certain level when the stopping criterion of the adaptive learning scheme is fulfilled. 

\begin{table}[!h]
\centering
\resizebox{\textwidth}{!}{%
\begin{tabular}{lllllll}
\hline
Stock price $S_1$ & Stock price $S_2$ & Exact option price & PIRBFNN 2200 & PIRBFNN 2600 & PIRBFNN 3000 & PIRBFNN 3400 \\ \hline
20 & 0 & 20.0000 & 19.9945 & 19.9947 & 19.9983 & 19.9990 \\
20 & 4 & 16.0000 & 15.9987 & 15.9957 & 15.9991 & 15.9996 \\
20 & 8 & 12.0000 & 12.0081 & 12.0049 & 12.0004 & 11.9998 \\
20 & 12 & 8.0052 & 8.0024 & 8.0117 & 8.0110 & 8.0070 \\
20 & 16 & 4.2372 & 4.2375 & 4.2405 & 4.2405 & 4.2401 \\
20 & 20 & 1.5931 & 1.6062 & 1.5935 & 1.5896 & 1.5900 \\
20 & 24 & 0.4295 & 0.4297 & 0.4276 & 0.4250 & 0.4217 \\
20 & 28 & 0.0900 & 0.0953 & 0.0910 & 0.0906 & 0.0903 \\
20 & 32 & 0.0159 & 0.0199 & 0.0201 & 0.0196 & 0.0181 \\
20 & 36 & 0.0025 & 0.0047 & 0.0056 & 0.0034 & 0.0016 \\
20 & 40 & 0.0004 & -0.0101 & 0.0007 & 0.0120 & 0.0074 \\
\hline
RMSE &  &  & 0.0063 & 0.0037 & 0.0046 & 0.0035
\end{tabular}%
}
\caption{Example 2: The numerical results for various asset prices $S_1$ and $S_2$ at time $t = 0$ obtained from PIRBFNNs with different numbers of RBF neurons during the adaptive learning process.}
\label{tab:Exchoptionvalue}
\end{table}

\begin{figure}[!h]
	\centering
    \subfigure[]{
    \includegraphics[width=2.5in]{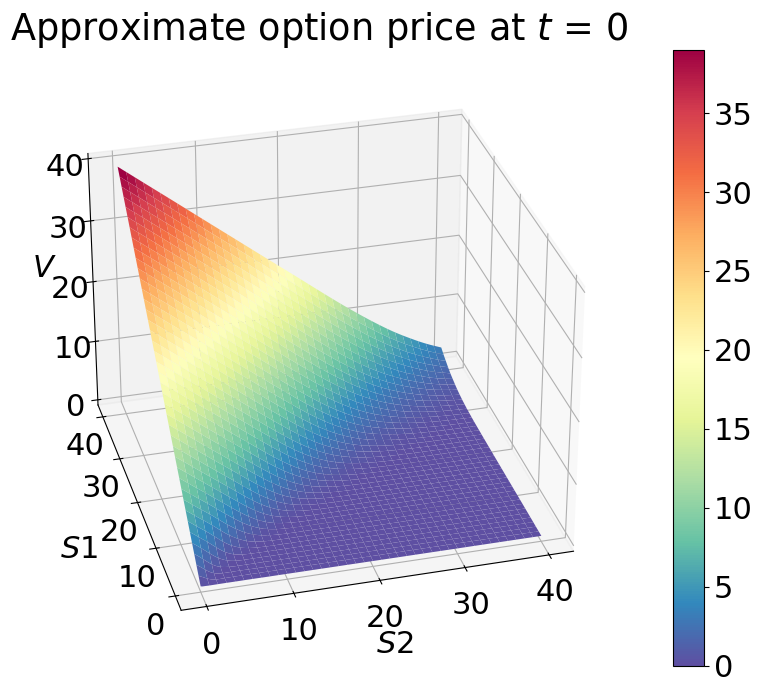}}%
    \subfigure[]{
    \includegraphics[width=2.5in]{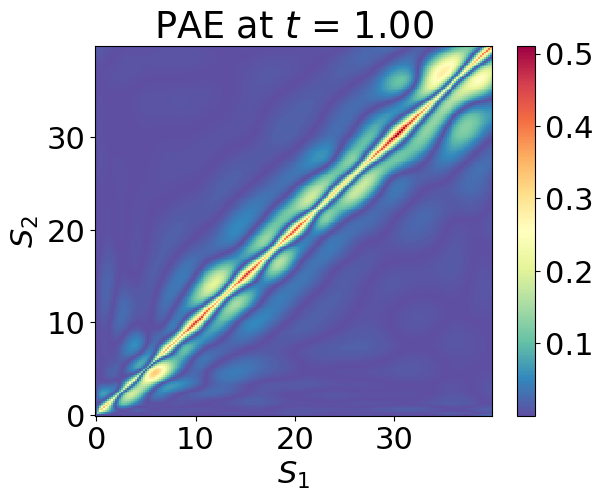}}
    \\
    \subfigure[]{
    \includegraphics[width=2.5in]{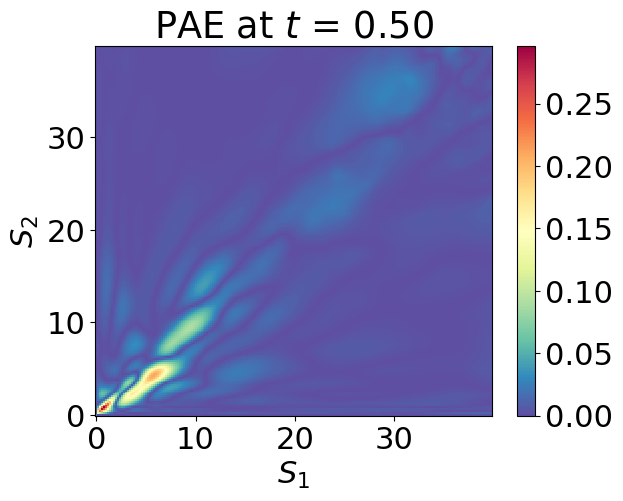}}%
    \subfigure[]{
    \includegraphics[width=2.5in]{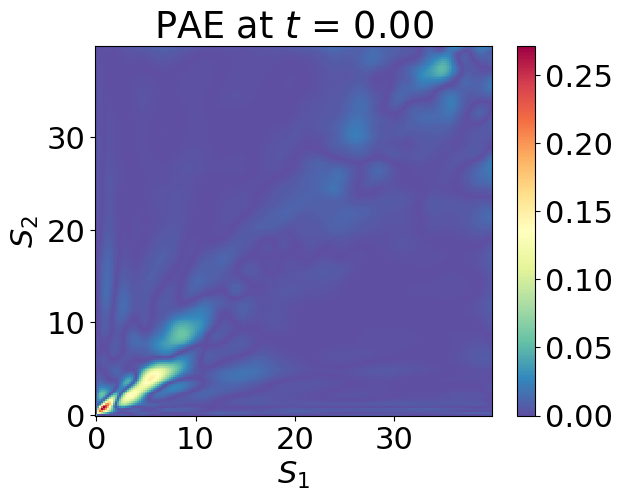}}
	\caption{Example 2: (a) The approximate solution profile of double-asset exchange option price at $t=0$ and the PAE contours at (b) $t = 1$, (c) $t = 0.5$, (d) $t = 0$.}
	\label{fig:Exchoptionprofile}
\end{figure}

An approximate solution profile of the exchange option zoomed in for $S_1$ values ranging from $0$ to $40$ and $S_2$ values ranging from $0$ to $40$ at $t = 0$ is depicted in the Fig.\ref{fig:Exchoptionprofile} (a), which is produced by the well-trained PIRBFNN after completing the adaptive learning process described above. As shown in Fig.\ref{fig:Exchoptionprofile} (b),(c), and (d), the largest error appears in the non-smooth region of the payoff function at the terminal time. However, our method achieves higher computational accuracy as time approaches $0$ without requiring pre-smoothing technique.

\subsection{Four-asset basket call option}
A basket option is a multi-asset financial contract that provides the holder with a payoff determined by the collective performance of multiple underlying assets. In a call option scenario, if the weighted sum of the underlying assets’s prices at expiration exceeds the strike price, the option pays out the difference between the two values; otherwise, it expires worthless. In contrast, in a put option scenario, the specified payoff materializes solely when the weighted sum of the underlying assets’ prices drops below the strike price. Given this context, a basket call option pricing problem involving four underlying assets can be addressed by solving the governing Black-Scholes Eq.\eqref{BS_equation_general} in the following manner
\begin{equation}\label{4DB-S}
	\begin{array}{l}
		\frac{\partial V(S_1,S_2,S_3,S_4,t)}{\partial t} + \frac{1}{2} \sigma_1^2 S_1^2 \frac{\partial^2 V(S_1,S_2,S_3,S_4,t)}{\partial S_1^2} + \frac{1}{2} \sigma_2^2 S_2^2 \frac{\partial^2 V(S_1,S_2,S_3,S_4,t)}{\partial S_2^2} + \frac{1}{2} \sigma_3^2 S_3^2 \frac{\partial^2 V(S_1,S_2,S_3,S_4,t)}{\partial S_3^2} + \\        \frac{1}{2} \sigma_4^2 S_4^2 \frac{\partial^2 V(S_1,S_2,S_3,S_4,t)}{\partial S_4^2} + \rho_{12} \sigma_1 \sigma_2 S_1 S_2 \frac{\partial^2 V(S_1,S_2,S_3,S_4,t)}{\partial S_1 \partial S_2} + \rho_{13} \sigma_1 \sigma_3 S_1 S_3 \frac{\partial^2 V(S_1,S_2,S_3,S_4,t)}{\partial S_1 \partial S_3} + \\ 
		\rho_{14} \sigma_1 \sigma_4 S_1 S_4 \frac{\partial^2 V(S_1,S_2,S_3,S_4,t)}{\partial S_1 \partial S_4} +
		\rho_{23} \sigma_2 \sigma_3 S_2 S_3 \frac{\partial^2 V(S_1,S_2,S_3,S_4,t)}{\partial S_2 \partial S_3} + \rho_{24} \sigma_2 \sigma_4 S_2 S_4 \frac{\partial^2 V(S_1,S_2,S_3,S_4,t)}{\partial S_2 \partial S_4} + \\
		\rho_{34} \sigma_3 \sigma_4 S_3 S_4 \frac{\partial^2 V(S_1,S_2,S_3,S_4,t)}{\partial S_3 \partial S_4} +
		 r S_1 \frac{\partial V(S_1,S_2,S_3,S_4,t)}{\partial S_1} +  r S_2 \frac{\partial V(S_1,S_2,S_3,S_4,t)}{\partial S_2} +  r S_3 \frac{\partial V(S_1,S_2,S_3,S_4,t)}{\partial S_3} + \\ 
		 r S_4 \frac{\partial V(S_1,S_2,S_3,S_4,t)}{\partial S_4} - r V(S_1,S_2,S_3,S_4,t) = 0,\\
		(S_1,S_2,S_3,S_4,t) \in [0,S_{max}]^4 \times [0,T],
	\end{array}
\end{equation}
equipped with the terminal payoff condition
\begin{equation}\label{4Dpayoff}
	V(S_1, S_2, S_3, S_4, T) = max(\sum_{i=1}^{4} \alpha_i S_i - K,0),
\end{equation}
where $\alpha_i (i = 1, 2, 3, 4)$ represents the proportion of the $i$-th asset within the basket of assets,
and the spatial boundary conditions
\begin{equation}\label{2Dfarboundary}
	\begin{cases} 
		V(S_{max},S_2,S_3,S_4,t) = max(\sum\limits_{i=1}^{4} \alpha_i S_i - K\cdot e^{- r (T-t)},0), \\[15pt]   
        V(S_1,S_{max},S_3, S_4,t) =max(\sum\limits_{i=1}^{4} \alpha_i S_i - K\cdot e^{- r (T-t)},0),  \\[15pt]   
        V(S_1, S_2,S_{max},S_4,t) = max(\sum\limits_{i=1}^{4} \alpha_i S_i - K\cdot e^{- r (T-t)},0),  \\[15pt] 
        V(S_1, S_2, S_3,S_{max},t) = max(\sum\limits_{i=1}^{4} \alpha_i S_i - K\cdot e^{- r (T-t)},0),  \\[15pt] 
		V(0,0,0,0,t) = 0.  
	\end{cases} 
\end{equation}

For simplicity, the domain we have used here is a hypercube. However, as demonstrated in \cite{PETTERSSON200882}, owing to the meshfree nature of the underlying approximation, it is possible to reduce the spatial domain to a simplex without loss of accuracy. This reduction enables significantly enhanced approximation performance in the region of interest while using the same number of neurons.

The numerical study is conducted employing the subsequent
model parameter values: $\sigma_1 = 0.4, \sigma_2 = 0.25, \sigma_3 = 0.3, \sigma_4 = 0.4$, $r = 0.05$,  $\rho_{11} = \rho_{22} = \rho_{33} = \rho_{44} = 1, \rho_{12} = \rho_{21} = 0.1, \rho_{13} = \rho_{31} = -0.4, \rho_{14} = \rho_{41} = 0.2, \rho_{23} = \rho_{32} = 0.3, \rho_{24} = \rho_{42} = -0.1, \rho_{34} = \rho_{43} = 0$, $\alpha_1=\frac{1}{4}$, $\alpha_2=\frac{1}{4}$, $\alpha_3=\frac{1}{4}$, $\alpha_4=\frac{1}{4}$, $T = 1$, $K = 1$, and $S_{max} = 4$ \cite{Tysk,Ficchera}. Here the choice of $S_{max} = 4$ is a trade-off that ensures high accuracy of the asymptotic far-field boundary conditions along the diagonal of the spatial domain; however, in the locations where $S_i$ are small, $i \neq j$, and $S_j=S_{max}$, the far-field boundary is too close to the strike price to be fully accurate.

For training the PIRBFNN, we employ a total of $M = 13600$ collocation points, consisting of $M_\mathcal{L} = 850 *10 =8500$ interior points  (where the factor $10$ corresponds to the number of boundaries of the five-dimensional spatial-temporal domain), $M_T = 850$ terminal time boundary points, and $M_\mathcal{B} = 850*5=4250$ spatial boundary points (with the factor $5$ representing the five spatial boundary conditions of a four-asset basket call option). Moreover, the learning rate of the L-BFGS optimizer is selected to be $0.5$.

By implementing the adaptive learning strategy, we initiate the training of a PIRBFNN with $N = 3000$ Gaussian RBF neurons in the hidden layer. Following this, at every $k = 200$th iteration before halting the training process, the complexity of the network is incrementally enhanced by adding $m = 200$ new neuron centres captured from those corresponding to the largest PDE residuals calculated at $s = 1500$ randomly selected checking points within the solution domain. At the end of each iteration $\tau$, we evaluate the training progress, which is concluded based on the change in the value of $\overline{\text{DWR}}_w^{(\tau)}$ with $w = 128$ and peak height threshold $\epsilon = 10^{-7}$.

\begin{figure}[!ht]
	\centering
    \includegraphics[width=0.45\textwidth]{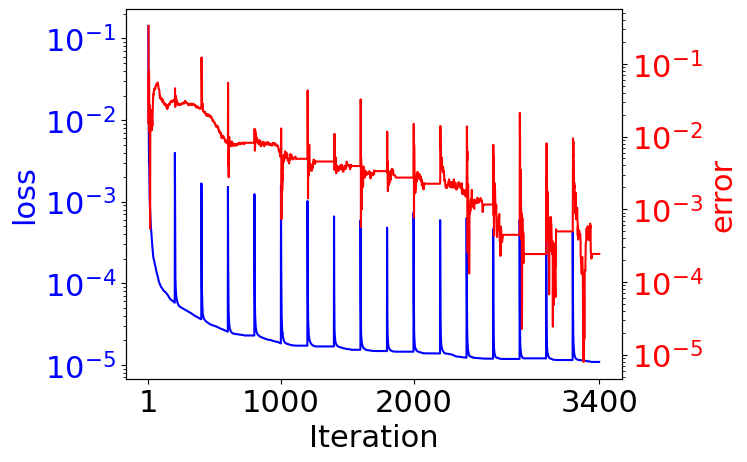}	
	\caption{Example 3: Adaptive training of the PIRBFNN for pricing four-asset basket call option: loss value history (blue curve), absolute error history (red curve).}
	\label{fig:4DRBFneurons}
\end{figure}

Due to the absence of an analytical solution for this model, we use the option value of $0.0971$ at time $t = 0$ with asset prices $(S_1, S_2, S_3, S_4) = (1, 1, 1, 1)$ provided by the Monte Carlo method as a reference to evaluate the accuracy of our network prediction. This target point $(S_1, S_2, S_3, S_4) = (1, 1, 1, 1)$ is located close to the bend of the solution, where numerical computation errors are typically more significant. After completing the adaptive training of the PIRBFNN, Fig.~\ref{fig:4DRBFneurons} illustrates the progression of the loss values and the absolute errors between the reference solution and the network approximation during the learning process, depicted by the blue and red curves, respectively. The results reveal that the PIRBFNN optimized via the adaptive hidden-layer expansion strategy, is capable of effectively addressing even a five-dimensional PDE problem, with time $t$ considered as an additional spatial variable. Nevertheless, due to the reliance on a single-provided evaluation from the reference solver as our testing information, the observed error trend is not always consistent with the declining trend of the training loss. Remarkably, applying the proposed learning algorithm to train the network until convergence, the network prediction at the testing point aligns well with the reference data, stably yielding a result of $0.0973$ with the absolute error reduced to $0.0002$ after $3400$ iterations.

To further assess the reliability of the adaptively trained PIRBFNN at additional locations, we sampled eight test points in the vicinity of the point $(1, 1, 1, 1)$. The approximation quality of the network at these points was also evaluated against reference solutions obtained via the Monte Carlo method. Using the same hyperparameters $(k, m,  s, w, \epsilon)$ as the previous experiment for adaptive training, Table \ref{tab:basoptionvalue} presents the predicted option values at $t = 0$ for the selected points, obtained using the trained PIRBFNN with different initial number of RBF neurons $N$ in their network architectures. The simulation results demonstrate that the adaptively trained PIRBFNN still maintains excellent performance at the newly sampled points. Notably, for the challenging four-asset option pricing model, the error achieves an order of $10^{-3}$ after training. Furthermore, it can be observed that prediction accuracy exhibits a pretty stability with the increasing number of initial neurons. Grounded in the experimental findings, it is also advisable to initiate training with a compact network architecture for the adaptive trained PIRBFNN. Leveraging the learning strategy proposed in this study, the network can be incrementally expanded during training, thereby ensuring accurate and stable predictive capability upon convergence as defined by the stopping criteria.

\begin{table}[]
\centering
\scalebox{0.9}{\begin{tabular}{lllllll}
\hline
\begin{tabular}[c]{@{}l@{}}Stock prices \\ $(S_1,S_2,S_3,S_4)$\end{tabular} &
  \begin{tabular}[c]{@{}l@{}}Reference \\ option prices\end{tabular} &
  \begin{tabular}[c]{@{}l@{}}PIRBFNN \\ $N=1500$\end{tabular} &
  \begin{tabular}[c]{@{}l@{}}PIRBFNN \\ $N=2000$\end{tabular} &
  \begin{tabular}[c]{@{}l@{}}PIRBFNN \\ $N=2500$\end{tabular} &
  \begin{tabular}[c]{@{}l@{}}PIRBFNN \\ $N=3000$\end{tabular} &
  \begin{tabular}[c]{@{}l@{}}PIRBFNN \\ $N=3500$\end{tabular} \\ \hline
(1.1,1.0,1.0,1.0) & 0.1142 & 0.1170 & 0.1159 & 0.1169 & 0.1166 & 0.1166 \\
(0.9,1.0,1.0,1.0) & 0.0814 & 0.0789 & 0.0780 & 0.0784 & 0.0781 & 0.0787 \\
(1.0,1.1,1.0,1.0) & 0.1133 & 0.1144 & 0.1137 & 0.1146 & 0.1136 & 0.1142 \\
(1.0,0.9,1.0,1.0) & 0.0823 & 0.0819 & 0.0807 & 0.0812 & 0.0816 & 0.0816 \\
(1.0,1.0,1.1,1.0) & 0.1131 & 0.1128 & 0.1118 & 0.1125 & 0.1120 & 0.1126 \\
(1.0,1.0,0.9,1.0) & 0.0826 & 0.0839 & 0.0829 & 0.0837 & 0.0835 & 0.0836 \\
(1.0,1.0,1.0,1.1) & 0.1145 & 0.1143 & 0.1135 & 0.1137 & 0.1133 & 0.1139 \\
(1.0,1.0,1.0,0.9) & 0.0811 & 0.0822 & 0.0809 & 0.0823 & 0.0820 & 0.0820 \\ \hline
RMSE                  &        & 0.0015 & 0.0016 & 0.0017 & 0.0016 & 0.0014 \\ \hline
\end{tabular}}
\caption{Example 3: The numerical results at time $t=0$ for various asset prices $(S_1, S_2, S_3, S_4)$ from PIRBFNNs initialized with varying numbers of RBF neurons within the adaptive training framework.}
\label{tab:basoptionvalue}
\end{table}

\begin{figure}[!ht]
	\centering
	\subfigure[]{
		\includegraphics[width=0.4\textwidth]{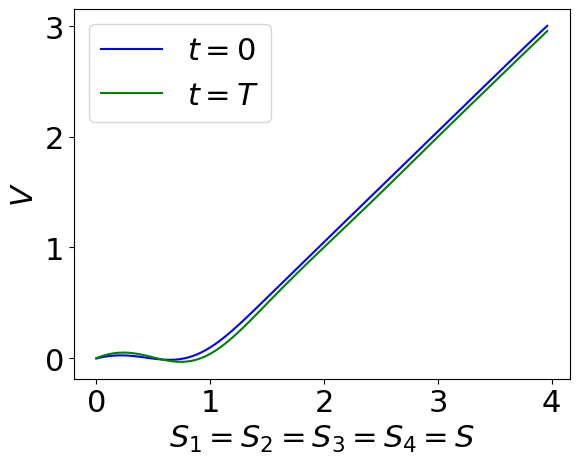}}
	\caption{Example 3: The approximate solution profiles of four-asset basket call option with asset prices $(S_1, S_2, S_3, S_4) = (S, S, S, S)$ when $t=T$ and $t=0$, respectively.}
	\label{fig:basoptionprofile}
\end{figure}

When the prices of all four underlying assets are set to the same value $S$ ranging from $0$ to $4$, Fig.~\ref{fig:basoptionprofile} illustrates the price evolution curves of the four-asset basket call option in relation to $S$ when time $t=T$ and $t=0$, which are predicted by the PIRBFNN with the best performance reported in Table \ref{tab:basoptionvalue}. The green curve represents $t=T$, while the blue curve corresponds to $t=0$. As can be observed from this figure, the solution values behave as expected in the strike region and for larger values of $S$; however, near $S=0$, both the solution and the initial condition exhibit small spurious oscillations.

\section{Conclusions}
In this study, the proposed physics-informed radial basis function neural network (PIRBFNN) can be regarded as a generalized radial basis function (RBF) meshless numerical method, while possessing the powerful machine learning capability of encoding the physics-informed (PI) laws described by the Black-Scholes PDE model. We designed the network architecture by incorporating the trainable centre locations, shape parameters, and linear coefficients for the typical RBF neurons in the hidden layer. Meanwhile, the collocation training points distribution, loss function formulation, and network parameter initialization are clearly demonstrated. Additionally, we conducted the learning dynamics of the PIRBFNN utilizing a second-order gradient decent optimizer and adaptively expanded the hidden layer based on the PDE residual information during the training process to further enhance prediction accuracy and training efficiency. Given the robustness of the machine learning mechanism, which minimizes the loss function through an iterative algorithm to optimize approximation quality, the PIRBFNN enables the adaptive distribution of centre points for the RBF neurons in the computational domain where they are most beneficial through the implementation of the proposed learning scheme. This approach avoids any ill-conditioning issues compromising stability which is likely to happen in the traditional RBF collocation methods. The numerical results showcase that the well trained PIRBFNNs are feasible in solving multi-dimensional option pricing problems. Throughout the experimental analysis, we have thoroughly examined the impact of various activation functions, iteration numbers, and neuron quantities on network performance. However, the presented PIRBFNN is confined to a single hidden layer configuration. Exploring the proper incorporation of multiple hidden layers into the current architecture could be a valuable avenue for future research to leverage the benefits of deep learning.

\section*{Acknowledgments}
The work of this paper was supported by National Natural Science Foundation of China (Grant NO.12161082) and Opening Project of Guangdong Province Key Laboratory of Computational Science at the Sun Yat-sen University (Grant NO.2024004).

\section*{Declarations of interest statement}
The authors declare that they have no competing interests.

\section*{Appendix A. Example 1: Investigation about the initial strategy for adaptive trained PIRBFNN with scalar shape parameters}

\begin{figure}[!h]
	\centering
    \subfigure[]{
	\includegraphics[width=2.5in]{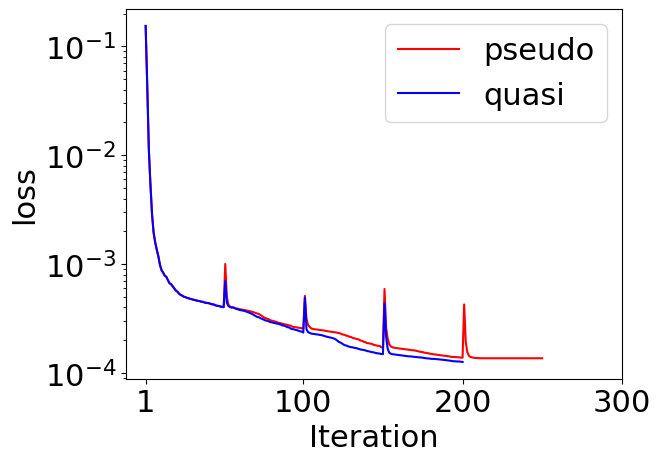}}
    \subfigure[]{
	\includegraphics[width=2.5in]{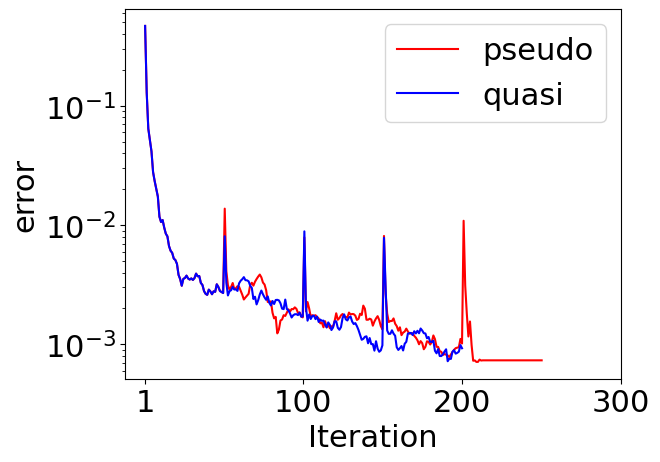}}
	\caption{Example 1: (a) loss value histories and (b) RMSE value histories for two initial centre points distributions: pseudo-random (red curve) and quasi-random (blue curve).} 
	\label{fig:halton}
\end{figure}

From the perspective of traditional methods to understand RBF, it is widely acknowledged that the location of centre points and the choice of shape parameters have a crucial impact on the performance of the algorithm. Therefore, in this supplementary part, we mainly focus on the influence of the initialization schemes with respect to centres distribution and shape parameters for PIRBFNN performance based on the linear weights still given by Xavier initialization.
For the model with $\sigma_1 = 0.2$, in the introduced adaptive algorithm, all other settings are kept unchanged except that the initial $350$ RBF neuron centre points are replaced with points sampled from a quasi-random (Halton) sequence instead of our proposed pseudo-random distribution (torch.rand). Additionally, in each iteration of the adaptive training, the set of $s=1000$ candidate points originally selected at random is now also sampled from the Halton sequence. This modification is made because these points may potentially be absorbed during adaptive training as the initial positions of newly added RBF neuron centres. The training loss and testing error curves of the PIRBFNNs, initialized with centre points using two different distribution strategies, are shown in Fig.\ref{fig:halton} (a) and (b), respectively. In this figure, the red curves correspond to the results obtained using pseudo-random point initialization, while the blue curves represent those obtained using quasi-random point initialization. At the end of adaptive training, the performance of the red and blue lines is comparable. With a pseudo-random arrangement of the initial centre points, the network converged in $250$ iterations, achieving an RMSE of $7.3520\times10^{-4}$ at the test points. In contrast, a quasi-random initialization required $200$ iterations to converge, with a test-point RMSE of $9.2824\times10^{-4}$.

\begin{figure}[!h]
	\centering
    \subfigure[]{
	\includegraphics[width=2.5in]{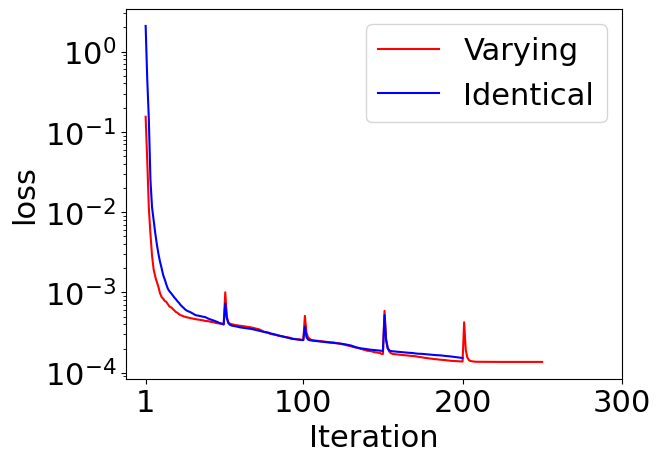}}
    \subfigure[]{
	\includegraphics[width=2.5in]{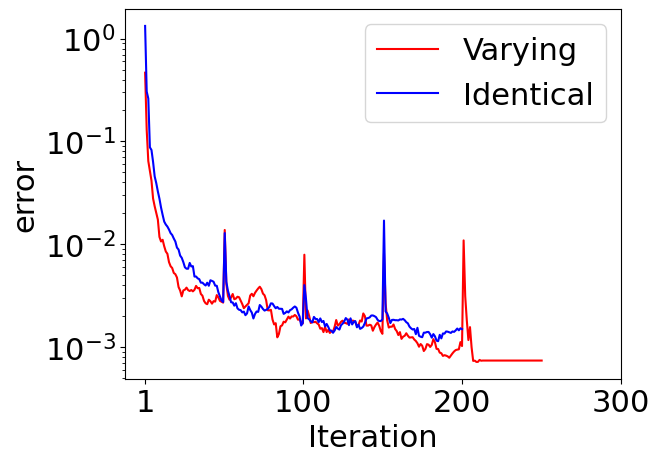}}
	\caption{Example 1: (a) loss value histories and (b) RMSE value histories for two initial shape parameter configurations: varying shape parameters (red curve) and identical shape parameter (blue curve).} 
	\label{fig:shape-1}
\end{figure}

Within the adaptive training framework of the PIRBFNN, although the difference in results between the two sampling styles for the initial centre points is small, the initialization strategy proposed in this study is still preferable. This is because it allows the use of random values for parameters, without requiring prior knowledge of the problem or careful selection of them. Subsequently, with the initialization approach suggested in this study for arranging the initial centre points, we set the shape parameter of each RBF neuron uniformly to a value of $1$ to start the learning. The training loss and testing error curves of the PIRBFNNs, corresponding to two different shape parameter initialization schemes, are shown in Fig.\ref{fig:shape-1} (a) and (b), respectively. In this figure, the red lines represent the results obtained using the proposed initialization strategy, in which different RBF neurons are assigned distinct shape parameters, whereas the blue lines represent the results obtained when all the initial RBF neurons share the same shape parameter. Although the blue curve ultimately achieves a slightly faster convergence rate, it exhibits lower approximation accuracy, with an RMSE of $1.4867\times10^{-3}$. Therefore, in all experiments conducted in this manuscript, we adopted the proposed initialization scheme during network training for accuracy as the primary concern.

\section*{Appendix B. Example 2: The behavior of the stop condition in the adaptive training for PIRBFNN with vectorized shape parameters}

\begin{figure}
	\centering
    \subfigure[]{\includegraphics[width=3in]{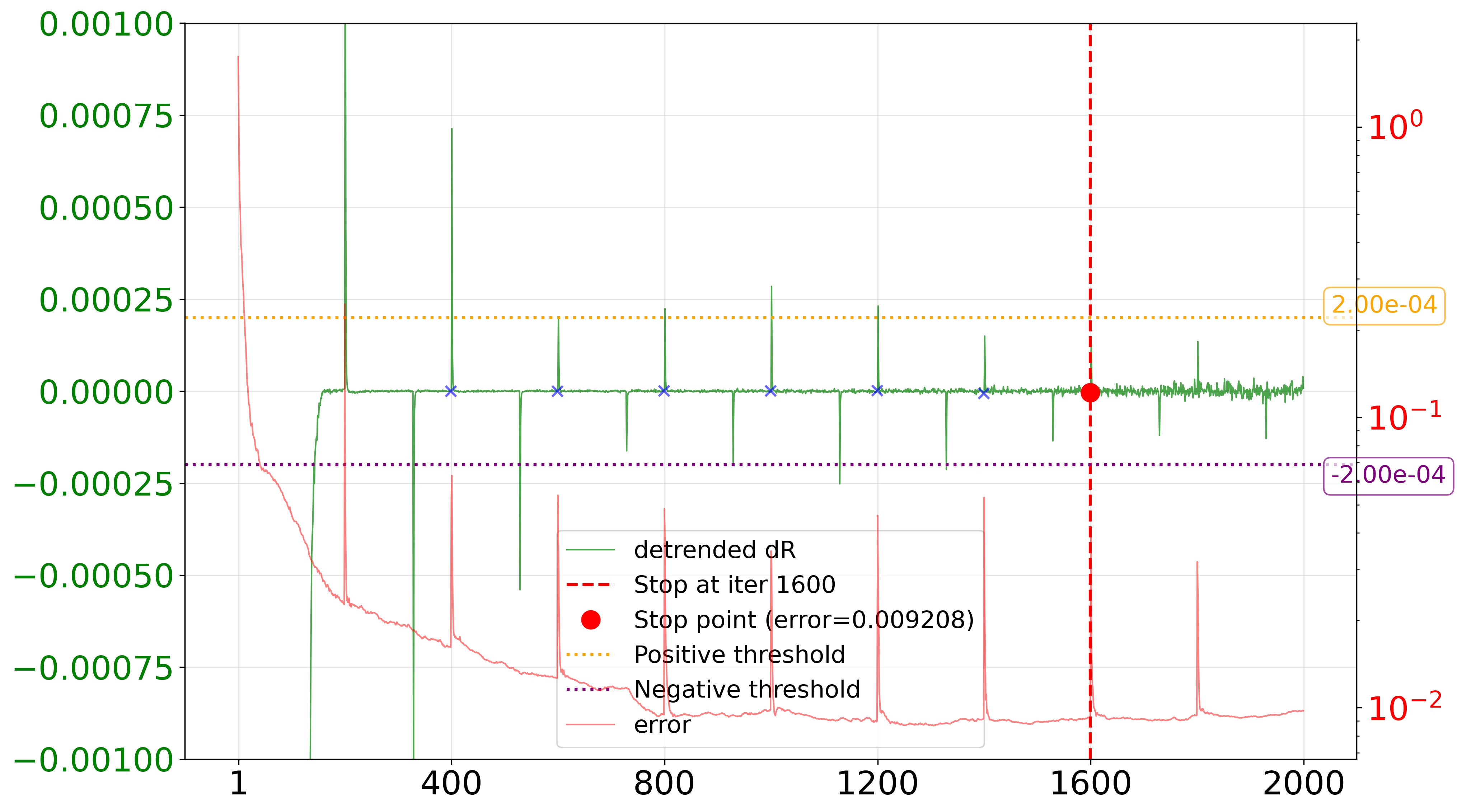}\label{fig:2d-stopCond-a}}
    \subfigure[]{\includegraphics[width=2.5in]{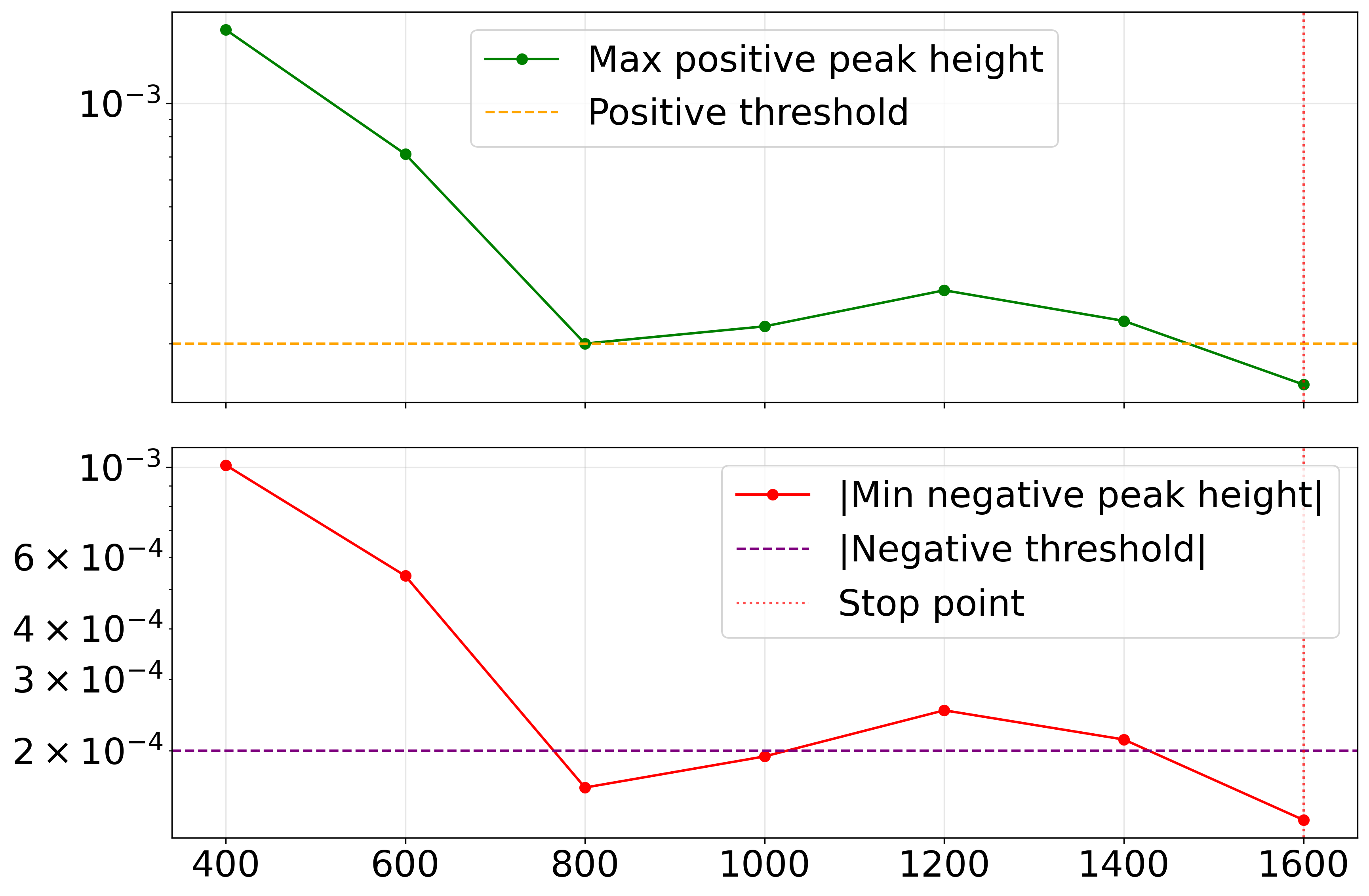}\label{fig:2d-stopCond-b}}
    \caption{Example 2: The behavior of the stop condition in the adaptive training.}
\end{figure}

We provide two figures that illustrate the interval-based stop condition used in Algorithm~\ref{alg:adaptive_training} on example 2. Similar phenomenons can be observed in other examples. The Fig.\ref{fig:2d-stopCond-a} plots the detrended residual-change signal $\bm{D}$ (i.e., $\bm{\Delta R}_w - \bm{T}$) over training iterations on the primary axis, with the testing error on a secondary axis in log scale. The blue markers indicate the iterations at which the stop condition is evaluated (every $k$ iterations), and the orange and purple horizontal lines show the positive and negative thresholds $\epsilon$ and $-\epsilon$. The iteration at which training is stopped is marked by a red vertical line and a point on the detrended signal. The Fig.\ref{fig:2d-stopCond-b} summarizes the peak heights at each evaluation step: the upper panel shows the maximum positive peak height $h_+$ versus the positive threshold (both in log scale), and the lower panel shows the absolute value of the minimum negative peak height $|h_-|$ versus the negative threshold. The vertical line again indicates the stop iteration. Together, these figures show how the detrended signal oscillates with decreasing amplitude and how the algorithm halts when both $h_+$ and $|h_-|$ fall below the chosen thresholds at an evaluation point. These figures thus motivate the use of the detrended residual-change peaks as a control signal: monitoring these peaks works better than monitoring the training loss, which always decreases even when overfitting causes the error to increase, and better than monitoring the raw residuals, which are highly noisy.

\section*{Appendix C. Example 3: Elapsed time versus error for adaptive PIRBFNN with varying initial RBF neuron counts}

\begin{figure}
	\centering
    \includegraphics[width=3in]{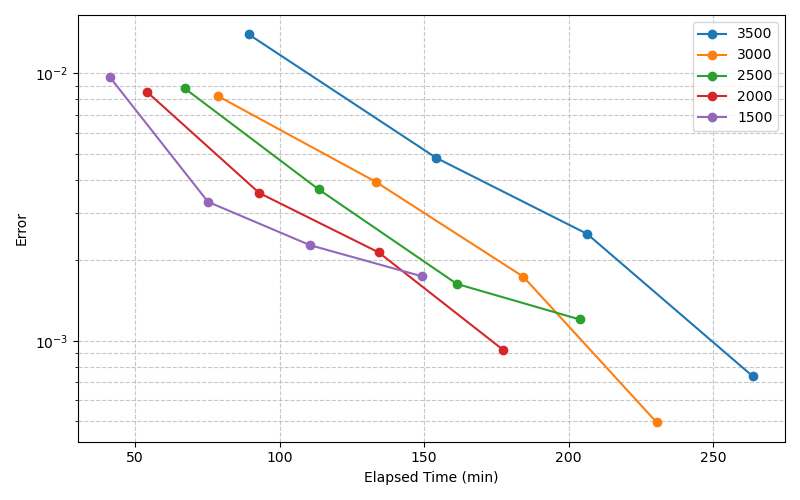}
    \caption{Example 3: Elapsed time versus error for adaptive PIRBFNN with varying initial RBF neuron counts.}
    \label{fig:4d-time}
\end{figure}

To illustrate the practical trade-offs when choosing the initial network size in the adaptive PIRBFNN framework for the four-asset basket call option, we compare five training runs that differ only in the initial number of RBF hidden neurons, $N \in \{1500, 2000, 2500, 3000, 3500\}$ in Fig. \ref{fig:4d-time}. It plots the testing error (vertical axis, log scale) against the elapsed wall-clock time in minutes (horizontal axis) at selected training iterations (e.g., 800, 1600, 2400, 3200). Each curve corresponds to one value of $N$; the detailed description on accuracy has been provided in Table \ref{tab:basoptionvalue}. The curves show that a larger initial $N$ typically incurs a higher computational cost per iteration (longer elapsed time to reach the same iteration count), but can yield lower errors along the training trajectory. This behaviour highlights a trade-off between upfront computational cost and achievable accuracy.

\bibliographystyle{elsarticle-num}
\bibliography{cited}

\end{document}